\documentclass[journal]{IEEEtran}
\usepackage{amsmath}
\usepackage{epsfig}
\usepackage{graphicx}
\usepackage{subfigure}
\usepackage{multirow}
\usepackage{color}
\usepackage{changepage}
\usepackage{times}
\usepackage{float}
\usepackage{amssymb}
\usepackage{multirow}
\usepackage{bbding}
\usepackage{booktabs}
\usepackage{algorithm}
\usepackage{algpseudocode}
\usepackage{caption2}
\usepackage{verbatim}
\usepackage{dsfont}
\usepackage{amssymb}
\usepackage{stfloats}
\usepackage{tikz}
\usetikzlibrary{arrows}

\begin{document}
\title{Learning a No-Reference Quality Assessment Model of Enhanced Images with Big Data}
\author{Ke~Gu,~
Dacheng~Tao,~
Junfei~Qiao,~
and~Weisi~Lin

\thanks{This work was supported in part by Australian Research Council Projects FT-130101457, DP-140102164, LP-150100671, Singapore MoE Tier 1 Project M4011379 and RG141/14.}
\thanks{K. Gu and J.-F. Qiao are with the Beijing Key Laboratory of Computational Intelligence and Intelligent System, BJUT Faculty of Information Technology, Beijing University of Technology, Beijing 100124, China (e-mail: guke.doctor@gmail.com; junfeiq@bjut.edu.cn).}
\thanks{D. Tao is with the School of Information Technologies and the Faculty of Engineering and Information Technologies, University of Sydney, J12/318 Cleveland St, Darlington NSW 2008, Australia (email: dacheng.tao@sydney.edu.au).}
\thanks{W. Lin is with the School of Computer Science and Engineering, Nanyang Technological University, Singapore, 639798 (email: wslin@ntu.edu.sg).}
\thanks{$\copyright$20XX IEEE. Personal use of this material is permitted. Permission from IEEE must be obtained for all other uses, in any current or future media, including reprinting/republishing this material for advertising or promotional purposes, creating new collective works, for resale or redistribution to servers or lists, or reuse of any copyrighted component of this work in other works.}}

\maketitle
\begin{abstract}
In this paper we investigate into the problem of image quality assessment (IQA) and enhancement via machine learning. This issue has long attracted a wide range of attention in computational intelligence and image processing communities, since, for many practical applications, e.g. object detection and recognition, raw images are usually needed to be appropriately enhanced to raise the visual quality (e.g. visibility and contrast). In fact, proper enhancement can noticeably improve the quality of input images, even better than originally captured images which are generally thought to be of the best quality. In this work, we present two most important contributions. The first contribution is to develop a new no-reference (NR) IQA model. Given an image, our quality measure first extracts 17 features through analysis of contrast, sharpness, brightness and more, and then yields a measre of visual quality using a regression module, which is learned with big-data training samples that are much bigger than the size of relevant image datasets. Results of experiments on nine datasets validate the superiority and efficiency of our blind metric compared with typical state-of-the-art full-, reduced- and no-reference IQA methods. The second contribution is that a robust image enhancement framework is established based on quality optimization. For an input image, by the guidance of the proposed NR-IQA measure, we conduct histogram modification to successively rectify image brightness and contrast to a proper level. Thorough tests demonstrate that our framework can well enhance natural images, low-contrast images, low-light images and dehazed images. The source code will be released at https://sites.google.com/site/guke198701/publications.
\end{abstract}
\IEEEpeerreviewmaketitle

\section{Introduction}
\IEEEPARstart{P}{hotos} captured via cameras/smart phones or created by computers always require post-processing towards better visualization and enhanced utility in various application scenarios, e.g. object detection and recognition. One of the main goals of such post-processing operations is to raise the image quality, such as visibility, contrast and brightness. Therefore, how to seek a well-designed image quality assessment (IQA) metric for faithfully predicting the quality of enhanced images, which can even optimize and improve enhancement methods, becomes a highly substantial and beneficial task.

Traditional IQA researches are mainly devoted to gauging commonly seen artifacts, for example, Gaussian blur, noise, JPEG/JPEG2000 compression, etc. One type of IQA studies is subjective assessment focusing on building image quality databases, e.g. LIVE [\ref{ref:01}], MDID2013 [\ref{ref:02}] and VDID2014 [\ref{ref:03}]. Via a carefully-prepared testing setting, the organizers invite sufficient inexperienced observers to rank testing images in a randomized presentation order, and then yield the final mean opinion scores (MOSs) by averaging all the valid observers' scores after some necessary post-processing procedures such as outliers screening. The other type of IQA explorations is concentrating on objective assessment. Typical objective IQA approaches are developed using mathematical models, neural networks [\ref{ref:41}] and learning systems [\ref{ref:05a}] to approximate real human judgements of image quality.

Subjective and objective assessments are both important and they play complementary roles. The former one provides benchmark results, which a good objective metric is expected to have a close correlation with. Yet subjective assessment usually costs dearly and consumes much time, and thus cannot be used in real-time and in-service systems. Resorting to the powerful computational ability of computers, objective metrics can serve to evaluate image quality in practical application scenarios, such as enhancement [\ref{ref:21}] and tone-mapping [\ref{ref:49}], replacing human beings to some extent.

The last few years have witnessed an explosive growth of objective visual quality assessment. Based on the accessibility of reference source images to be compared with during the experiments, objective IQA approaches can be classified into three categories, i.e. \emph{full-reference} (FR) IQA [\ref{ref:05a}, \ref{ref:05}, \ref{ref:06}, \ref{ref:08}, \ref{ref:09}, \ref{ref:09b}], \emph{reduced-reference} (RR) IQA [\ref{ref:09a}, \ref{ref:10}, \ref{ref:11}, \ref{ref:12}, \ref{ref:13}], and \emph{no-reference} (NR)/blind IQA [\ref{ref:14}, \ref{ref:17}, \ref{ref:15}, \ref{ref:19}, \ref{ref:15a}]. Using popular large-size image databases, e.g. LIVE, TID2008 [\ref{ref:24}], CSIQ [\ref{ref:25}] and TID2013 [\ref{ref:26}], most of the above-mentioned IQA models have been proved of fairly high performance in accordance with subjective assessment.

The majority of current blind IQA methods were proposed based on two steps, namely feature extraction and SVR-based regression module. In these NR-IQA algorithms, more efforts were made to explore more valid features towards simulating the perceptual characteristics of human eyes to estimate the visual quality. With considerable effective features developed, a growing body of researchers turn to resorting to advanced neural networks and learning systems, e.g. general regression neural network [\ref{ref:41}], multiple kernel learning [\ref{ref:16}], deep belief net [\ref{ref:18}, \ref{ref:18a}] and pairwise learning-to-rank approach [\ref{ref:20}], for the purpose of better approaching the ability of human eyes to group perceptual features and thereby more reliably inferring the overall quality score.

The majority of the IQA approaches described above are largely limited to commonly encountered artifacts. But with the development of compression, transmission and restoration technologies during last few decades, the above-mentioned artifacts might be not the leading factor of image quality any more. In comparison, IQA of enhancement very possibly plays a more significant role, since enhancement technologies are able to generate better images, even outperforming originally captured images which are usually thought to have the optimal quality. Unfortunately, the aforesaid IQA methods fail in this problem, because most of them directly or indirectly suppose that original natural images or the images that conform to statistics regulations observed from natural images [\ref{ref:37}] have the best quality and hence cannot correctly judge the quality of properly enhanced images [\ref{ref:27}].

\begin{figure}[!t]
\small
\centering
\vspace{0.05cm}
\hspace{-0.1cm}\subfigure[]{\setcounter{subfigure}{1}\includegraphics[height=2.25cm]{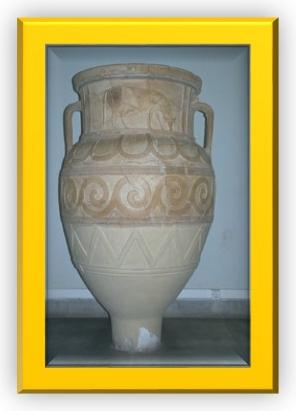}}
\hspace{-0.1cm}\subfigure[]{\setcounter{subfigure}{3}\includegraphics[height=2.25cm]{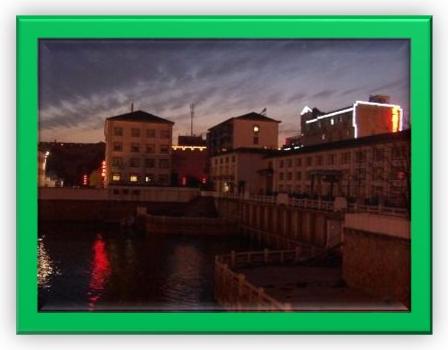}}
\hspace{-0.1cm}\subfigure[]{\setcounter{subfigure}{5}\includegraphics[height=2.25cm]{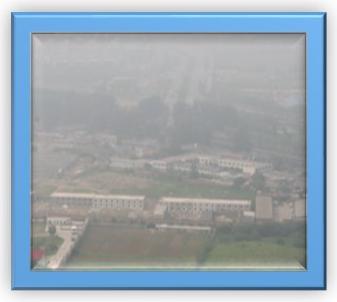}}
\hspace{-0.1cm}\subfigure[]{\setcounter{subfigure}{7}\includegraphics[height=2.25cm]{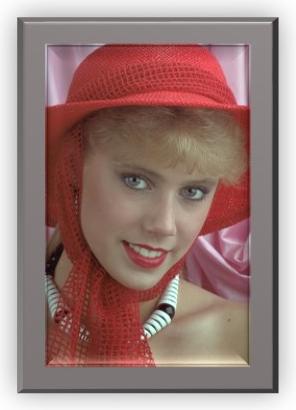}}\\
\hspace{-0.1cm}\subfigure[]{\setcounter{subfigure}{2}\includegraphics[height=2.25cm]{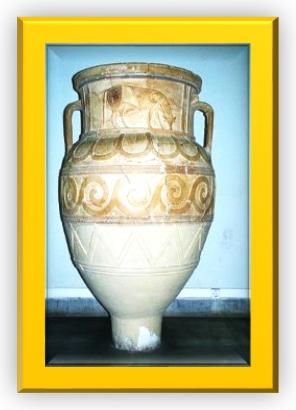}}
\hspace{-0.1cm}\subfigure[]{\setcounter{subfigure}{4}\includegraphics[height=2.25cm]{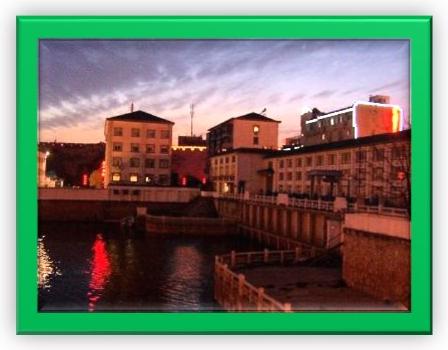}}
\hspace{-0.1cm}\subfigure[]{\setcounter{subfigure}{6}\includegraphics[height=2.25cm]{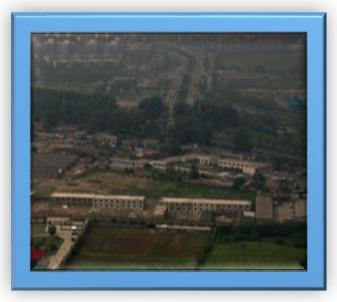}}
\hspace{-0.1cm}\subfigure[]{\setcounter{subfigure}{8}\includegraphics[height=2.25cm]{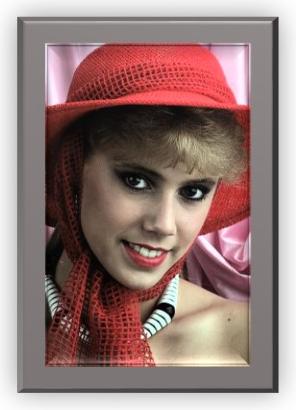}}\\
\vspace{-0.15cm}
\caption{\small Illustration of enhanced images: (a)-(b) natural image and its enhanced version [\ref{ref:27}]; (c)-(d) night image and its enhanced version [\ref{ref:28}]; (e)-(f) haze image and its dehazed image [\ref{ref:29}]; (g)-(h) natural image and its enhanced one by histogram equalization.}
\label{fig:01}
\end{figure}

Appropriate image enhancement technologies can raise the visual quality, as exemplified in Figs. \ref{fig:01}(a)-(f), while improper technologies degrade the quality, as shown in Figs. \ref{fig:01}(g)-(h). So accurately assessing the quality of enhanced images and judging the enhancement is proper or not have aroused much attention of researches during recent years. Gu \emph{et al.} first systematically studied this issue; they built up the CID2013 and CCID2014 databases dedicated to image contrast change, and meanwhile proposed RR-IQA techniques based on phase congruency and information statistics of the image histogram [\ref{ref:27}, \ref{ref:30}]. Another RR-IQA algorithm was devised by taking account of the fact that properly enhanced images should be simultaneously of entropy increment and saliency preservation [\ref{ref:31}]. Very lately, Wang \emph{et al.} put forward a FR quality metric by adaptively representing the structure of each local patch. To specify, this approach decomposes each image patch into three components, mean intensity, signal strength and signal structure, followed by separately measuring their perceptual distortions to be merged into one score [\ref{ref:32}].

As for most enhanced images, we are unable to obtain the associated original references. The aforesaid FR- and RR-IQA measures are unable to work in this situation, and therefore blind/NR algorithms are eagerly required. Not long ago, Fang \emph{et al.} proposed a dedicated blind quality metric based on the natural scene statistics (NSS) regulation, which involves mean, standard deviation, skewness, kurtosis and entropy [\ref{ref:33}]. One major limitation of this blind metric is that the natural images are considered to be of the highest quality. Also, this metric overlooks significant influences factors, e.g. colorfulness and local sharpness. In [\ref{ref:34}], Chen \emph{et al.} used a concatenation of GIST descriptor [\ref{ref:35}] and color motion [\ref{ref:36}] as 521-dimensions features before conducting a regression module to derive the final quality measure. Despite promising performance, using such high-dimension features easily introduces overfitting and there lacks definite connections and analyses between the used features and IQA of enhancement.

In this paper we propose a novel two-step framework for blind image quality measure of enhanced images (BIQME). Contrast is defined to be the difference in luminance or color that makes an object (or its representation in an image or display) distinguishable [\ref{ref:37a}]. Compared with luminance contrast which reflects the variations in luminance, color contrast also includes the variations in saturation and hue. Based on this concern, in the first step, we comprehensively consider five influencing factors which consist of contrast, sharpness, brightness, colorfulness and naturalness of images, and extract a total of 17 features. A high-quality image should have comparatively large contrast and sharpness, making more details highlighted. For these two types of features, we use modified entropy, contrast energy and log-energy of wavelet subbands. Besides, proper brightness and colorfulness usually render the whole image a broader dynamic range, which is beneficial to appear details as well. The last concern is the naturalness which a good-looking image is expected to be of. This work uses the classical NSS model [\ref{ref:37}] and recently released dark channel prior (DCP) [\ref{ref:29}] to estimate the naturalness of images. In the second step, we focus our attention on learning the regression module from extracted features above. Differing from current works which just use a small number of training data [\ref{ref:14}, \ref{ref:17}, \ref{ref:16}, \ref{ref:18}, \ref{ref:20}, \ref{ref:33}], we have gathered beyond 100,000 enhanced images (much larger than the size of related image databases) as big-data training samples and their corresponding objective quality scores derived by a newly designed high-accuracy FR-IQA model as training labels to learn the module of the proposed NR quality metric. There is no overlapping between the 100,000 training images and testing images in enhancement-related quality databases. Comparative tests confirm the superior performance and low computational cost of our measure relative to state-of-the-art FR-, RR- and NR-IQA methods. In view of the efficacy and efficiency, our IQA model severs as an optimization criterion to guide a histogram modification technology for enhancing images. The proposed enhancement method is shown to raise the visual quality of natural images, low-contrast images, low-light images and dehazed images.

In comparison to previous works, five contributions of this paper are summarized below: 1) to the best of our knowledge, this work is the first opinion-unaware\footnote{Generally, it needs training images labeled by subjective quality scores in opinion-aware metrics, while opinion-unaware methods do not require human scoring procedures and such human-labeled training images. Opinion-unaware metrics usually have more potential for good generalization ability.} blind IQA metric for image enhancement; 2) we establish a novel IQA framework from five influencing variables concerning enhancement; 3) a huge number of 100,000 training data are employed to build our BIQME metric, compared with only hundreds of training samples used in current NR-IQA models; 4) our blind metric performs better than most recently developed FR-, RR- and NR-IQA techniques on relevant image databases; 5) a new robust image enhancement technology is explored based on BIQME-optimization.

The remainder of this paper are organized as follows: In Section \ref{sec:2}, we propose the blind BIQME method as well as a modified FR IQA model. In Section \ref{sec:3}, thorough experiments verify the superiority and efficiency of our BIQME metric in contrast to modern FR-, RR- and NR-IQA measures. Section \ref{sec:4} presents the quality-optimized robust image enhancement approach. Section \ref{sec:5} concludes the whole paper.

\section{No-Reference Quality Metric}
\label{sec:2}
The design philosophy of our blind BIQME metric lies in five influencing factors, namely, contrast, sharpness, brightness, colorfulness and naturalness of images; the corresponding total 17 features are extracted accordingly. Afterwards, a regression module which is learned via a huge number of training data is used to fuse the aforementioned 17 features for inferring the ultimate quality score.

\begin{table*}[!t]
\vspace{-0.01cm}
\center
\caption{\small Summary of extracted features for IQA of enhancement.}
\label{tab:01}
\vspace{0.2cm}
\includegraphics[width=13.75cm]{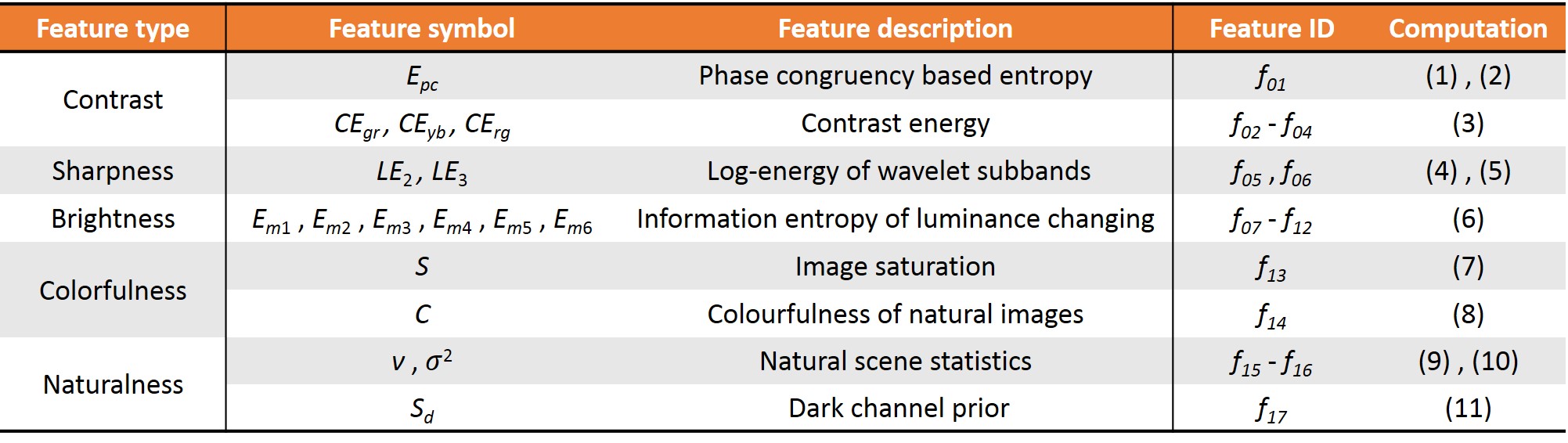}\\
\end{table*}

\subsection{Feature Extraction}
\label{sec:2.1}
Contrast is the leading factor which decides the effect of image enhancement. Information entropy is a classical and frequently used measurement of image contrast. Entropy is a global measurement, which characterizes the average amount of information contained in an image. In general, a greater entropy means that an image is of larger contrast and thereby of better visual quality. We take two images shown in Figs. \ref{fig:01}(c)-(d) as an example. It is quite obvious that image (c) with entropy 6.9 is visually worse than image (d) with entropy 7.6. Due to the limited processing ability, human brain is incline to pay attention to the regions which stores more perceptual information as priority. The phase congruence (PC) principle unveils that, as opposed to the Fourier amplitude, the Fourier phase contains higher amount of perceptual information [\ref{ref:38}]. Subsequently, it has been further demonstrated that mammals extracted features at the areas where the Fourier components are maximal in phase [\ref{ref:39}]. Hence we deploy a simple but biologically plausible PC model to detect and identify features in an image [\ref{ref:41}, \ref{ref:40}] and thus compute the PC-based entropy.

More specifically, similar to [\ref{ref:27}], we denote $M^o_n$ and $M^e_n$ filters which implement on scales $n$ with the odd- and even-symmetric properties. These two filters are constructed based on the log-Gabor function, because of its ability to maintain DC component and encode natural images [\ref{ref:06}]. In this work, we deploy the 2-D log-Gabor function defined by $G(\omega,o_k)=\exp[-\frac{(\log(\omega/\omega_0))^2}{2\sigma_r^2}]\cdot\exp[-\frac{(o-o_k)^2}{2\sigma_{o}^2}]$, where $o_k=k\pi/K$, $\omega$ is the center frequency of filters, $\sigma_r$ controls the bandwidth of filters, $k=\{0,1,...,K-1\}$ is the filter's orientation angle, $K$ is the number of orientations, and $\sigma_{o}$ decides the angular bandwidth of filters. By adjusting $\omega$ and $o_k$, we accordingly generate odd- and even-symmetric $M^o_n$ and $M^e_n$ filters, and further generate a quadrature pair for an image signal $\mathbf{s}$. At position $j$ on scale $n$, each quadrature pair is taken action to yield a response vector $[e_n(j),o_n(j)]$ = $[s(j)\ast M^e_n, s(j)\ast M^o_n]$, whose the amplitude value is $A_n(j)=\sqrt{e_n(j)^2+o_n(j)^2}$. Let $F(j)=\sum_ne_n(j)$ and $H(j)=\sum_no_n(j)$. PC is defined as $PC(j)=\frac{U(j)}{\varepsilon+\sum_nA_n(j)}$, where $U(j)=\sqrt{F^2(j)+H^2(j)}$ and $\varepsilon$ is a very small number to avoid division-by-zero. By simplification, PC can be computed by
\begin{equation}
PC(j)=\frac{\sum_nW(j)\lfloor A_n(j)\cdot\Delta\theta_n(j)-T_n\rfloor}{\varepsilon+\sum_nA_n(j)}
\label{func:01}
\end{equation}
where $\lfloor\rfloor$ is a threshold used to delete negative results through setting them to zero. $T_n$ predicts the noise extent. $\Delta\theta_n(j)=\cos[\theta_n(j)-\overline{\theta(j)}]-|\sin[\theta_n(j)-\overline{\theta(j)}]|$ is exploited to gauge the deviations in phase. $\overline{\theta(j)}$ is defined as the mean values of phase at $j$. $W(j)=(1+\exp[(u-t(j))v])^{-1}$ is manipulating function by weighting. $t(j)=\frac{1}{N}\sum_nA_n(j)(A_{\max}(j)+\varepsilon)^{-1}$. As for filter responses, $u$ offers a cut-off value for penalizing low PC values under it. $v$ is defined as a gain variable that control the cutoff sharpness. As thus, the PC-based entropy is defined by
\begin{equation}
E_{pc}=-\sum_{i=0}^{255} P_i(\mathbf{s}_{pc})\cdot\log P_i(\mathbf{s}_{pc})
\label{func:02}
\end{equation}
where $\mathbf{s}_{pc}$ is constituted by the pixels in $\mathbf{s}$, which corresponds to the 40\% largest values in the detected PC map.

The second measurement is contrast energy, which estimates perceived image local contrast [\ref{ref:42}]. The reason behind using it lies in that contrast energy has computational simplicity and particularly contrast-aware attributes [\ref{ref:43}]. We apply Gaussian second-order derivative filters to separate an image. The entire filter responses were adjusted with rectification and divisive normalization for modeling the process of nonlinear contrast gain control in visual cortex [\ref{ref:44}]. Similar to [\ref{ref:45}], we compute contrast energy on three channels:
\begin{equation}
CE_f=\frac{\alpha\cdot Y(\mathbf{s}_f)}{Y(\mathbf{s}_f)+\alpha\cdot\theta}-\phi_f
\label{func:03}
\end{equation}
where $Y(\mathbf{s}_f)=\sqrt{(\mathbf{s}_k*f_h)^2+(\mathbf{s}_k*f_v)^2}$. $f=\{gr,yb,rg\}$ are respectively three channels of $\mathbf{s}$, where $gr=0.299R+0.587G+0.114B$, $yb=0.5(R+G)-B$ and $rg=R-G$ [\ref{ref:45a}]. For parameters, $\alpha=\max[Y(\mathbf{s}_f)]$, $\theta$ governs the contrast gain, and $\phi_f$ is applied to constrain the noise with threshold. $f_h$ and $f_v$ separately stand for horizontal and vertical second-order derivatives of Gaussian function. Hence contrast-related features are defined as $\textit{F}_{\textit{ct}}=\{\textit{E}_{\textit{pc}},\textit{CE}_{\textit{gr}},\textit{CE}_{\textit{yb}},\textit{CE}_{\textit{rg}}\}$.

Sharpness is another influencing variable with comparable importance of image contrast. Contrary to contrast that fixes on the global sensation in our work, sharpness more perceives local variations. Intuitively speaking, for a photo, fine details are usually resolvable in sharp regions, such as edges and object boundaries. In application scenarios, many professional photographers try to alter perceived sharpness of a photo to a considerable high level. Typical solutions are composed of using high-resolution cameras and resorting to post-processing techniques such as retouching [\ref{ref:46}].

Actually, these years have seen quite a few works dedicated to sharpness assessment [\ref{ref:46a}, \ref{ref:47}, \ref{ref:48}]. According to [\ref{ref:47}], we choose an efficient and effective way to compute log-energy of wavelet subbands. To be more concretely, we first use 9/7 DWT filters to decompose a grayscale image into three levels, namely $\{LL_3, LH_l, HL_l, HH_l|l=1,2,3\}$. Considering the fact that more high-frequency details are generally contained in high-sharp images, we then compute the log-energy of each wavelet subband at each decomposition level to approximate this fact:
\begin{equation}
LE_{k,l}=\log_{10}\bigg[1+\frac{1}{K_l}\sum_ik_l^2(i)\bigg]
\label{func:04}
\end{equation}
where $i$ stands for the pixel index; $k$ is $LH$, $HL$, and $HH$, respectively; $K_l$ is the total number of DWT coefficients at the level $l$. Lastly, the log-energy at each decomposition level is calculated by
\begin{equation}
LE_{l}=\frac{\frac{1}{2}(LE_{LH,l}+LE_{HL,l})+w\cdot LE_{HH,l}}{1+w}
\label{func:05}
\end{equation}
where the parameter $w$ is assigned to be 4 to impose larger weights on $HH$ subbands. Here we merely take the 2nd and 3rd levels into consideration, since they involve more sharp details and results illustrate that adding the 1st level cannot result in performance gain in our BIQME model. Sharpness-related features are thus defined as $\textit{F}_{\textit{s}}=\{\textit{LE}_{2},\textit{LE}_{3}\}$.

Brightness highly affects the effect of image enhancement, since on one hand appropriate image brightness can render an image a broader dynamic range, and on the other hand it may contain semantic information, for example, providing scene information $-$ daylight seaside, dark-night seabed, and more. In this regard, we characterize image brightness with a simple strategy, following a recent work regarding IQA of tone-mapping operators [\ref{ref:49}]. Particularly, we hypothesize that proper brightness had better help images display more details, regardless of in dark regions or bright regions. That is to say, no matter whether holding, increasing or decreasing the luminance intensity, one good enhanced image is capable of preserving much information. By this guidance, we first create a set of intermediate images by raising/reducing the original brightness of an image
\begin{equation}
\mathbf{s}_{i}=\max(\min(m_i\cdot\mathbf{s},t_{u}),t_l)
\label{func:06}
\end{equation}
where $m_i$ indicates the multiplier index to be discussed later; $t_l$ and $t_u$ are the lower bound and upper bound; $\max$ and $\min$ are applied to restrain the image signal into range of $[t_l,t_u]$. In this paper, we temporarily only consider 8-bit images and therefore set $t_l$ and $t_u$ to be 0 and 255 respectively.

It is clear that, as the luminance intensity varies like this, image details will be removed. Hence we next compute how fast the details disappear. Various kinds of measurements can be leveraged in this work, such as mean, variance, entropy, nonsymmetric K-L divergence, symmetric J-S divergence, etc. According to some observations shown in [\ref{ref:49}], information entropy of the aforesaid intermediate images can effectively discriminate two photos that are captured in well-exposure and bad-exposure (including over-exposure and under-exposure) conditions, respectively. Indeed, even as for two properly exposed photos, this strategy also takes effect to judge their relative quality. Accordingly this paper deploys entropy of luminance-varying images to deduce whether an image has suitable brightness or not. Facing the choice of multiplier index $m_i$, more indices are beneficial to give rise to greater performance yet do harm to computation speed. So we find a good balance between efficacy and efficiency by just using six entropy values $\{E_{m1},E_{m2},...,E_{m6}\}$, which are measured with $m=\{n,\frac{1}{n}|n=3.5,5.5,7.5\}$. It deserves emphasis that, different from [\ref{ref:49}], we do not include entropy of the image $\mathbf{s}$ itself, because a similar measure $E_{pc}$ has been taken into consideration. As stated above, we define brightness-related features as $\textit{F}_{\textit{b}}=\{\textit{E}_{\textit{m}1},\textit{E}_{\textit{m}2},\textit{E}_{\textit{m}3},\textit{E}_{\textit{m}4},\textit{E}_{\textit{m}5},\textit{E}_{\textit{m}6}\}$.

Colorfulness has an akin function of brightness, offering a color image with wider dynamic range and thereby showing more details and information relative to a grayscale image. To quantify image colorfulness, we first introduce color saturation, which represents the colorfulness of a color compared with its own luminance. Here we simply compute the global mean of saturation channel after transforming an image into the HSV color space
\begin{equation}
S=\frac{1}{M}\sum_{i=1}^MT_{X\rightarrow S}[\mathbf{s}(i)]
\label{func:07}
\end{equation}
where $T_{X\rightarrow S}$ stands for a transformation function to convert an $X$ type image (e.g. RGB image) into the saturation channel; $M$ indicates the number of pixels in $\mathbf{s}$.

The second measurement stems from a classical research dedicated to measuring colourfulness in natural images [\ref{ref:45a}]. In fact, several well-designed colour appearance models can predict the perception of colourfulness, but they just work validly for simple blocks on a uniform background. As for the measurement of the global colourfulness of natural scene images, there is still no particular study. Through key features extraction and a psychophysical category scaling experiment, Hasler \emph{et al.} have contributed a practical metric to estimate the overall image colourfulness, which highly correlates with human perceptions [\ref{ref:45a}]. More detailedly, four key features are first extracted, consisting of the mean and variance of $yb$ and $rg$ channel ($\mu_{yb}$, $\sigma_{yb}^2$, $\mu_{rg}$ and $\sigma_{rg}^2$). Then the metric is defined by
\begin{equation}
C=\sqrt{\sigma_{yb}^2+\sigma_{rg}^2}+\kappa\sqrt{\mu_{yb}^2+\mu_{rg}^2}
\label{func:08}
\end{equation}
where $\kappa$ is a parameter to rectify the relative significance, in order to match subjective opinions better. Experimental results show that the optimal value of $\kappa$ is 0.3. Colorfulness-related features are therefore defined as $\textit{F}_{\textit{cl}}=\{\textit{S},\textit{C}\}$.

Naturalness is the intrinsic attribute of an natural image, which presents some commonness of the majority of natural images, e.g. the NSS regulation applied in [\ref{ref:14}, \ref{ref:17}]. Generally speaking, violating this regulation means that an image looks unnatural and thus is of low visual quality. Nonetheless, as mentioned above, a natural image will acquire better quality via proper enhancement. So the use of image naturalness is mainly to punish over-enhancement conditions, which usually seriously devastate the naturalness of a visual signal. Our first consideration is the typical and frequently used NSS model [\ref{ref:14}, \ref{ref:17}, \ref{ref:37}]. Specifically, we begins by preprocessing an image via local mean removal and divisive normalization:
\begin{equation}
\mathbf{s}(i)^*=\frac{\mathbf{s}(i)-\mu(i)}{\sigma(i)+\epsilon}
\label{func:09}
\end{equation}
where $\mu(i)$ and $\sigma(i)$ are local mean and standard deviation at the $i$-th pixel; $\epsilon$ is a positive constant. Then, as for a natural image, the normalized pixel values tend towards a Gaussian-like appearance, while the artifacts change the shape, for instance, Gaussian blur generates a more Laplacian appearance. The generalized Gaussian distribution (GGD) with zero mean was found to catch the behavior of coefficients of (\ref{func:09}), which is defined by
\begin{equation}
f(x;\nu,\sigma^2)=\frac{\nu}{2\beta\Gamma(\frac{1}{\alpha})}\exp\bigg(-\bigg(\frac{|x|}{\beta}\bigg)^\nu\bigg)
\label{func:10}
\end{equation}
where $\beta=\sigma\sqrt{\frac{\Gamma(\frac{1}{\nu})}{\Gamma(\frac{3}{\nu})}}$ and $\Gamma(a)=\int_0^{\infty}t^{a-1}e^{-t}dt$ when $a>0$. The parameter $\nu$ controls the \textit{shape} of the distribution while $\sigma^2$ means the \textit{variance} of the distribution. We therefore collect $\nu$ and $\sigma^2$ as two features.

The other measurement of naturalness is the recently found DCP prior [\ref{ref:29}], in which it shows that, in most non-sky areas, at least one color channel tend towards zero, that is
\begin{equation}
\mathbf{s}_{\textrm{dark}}(i)=\min_{k\in\{R,G,B\}}\mathbf{s}_{k}(i)
\label{func:11}
\end{equation}
where $k=\{R,G,B\}$ means the RGB channels. Apparently, $\mathbf{s}_{\textrm{dark}}$ has definite bounds of $[0,255]$ or $[0,1]$ for a normalized image divided by 255. We merely compute the overall mean of the dark channel $\mathbf{s}_{\textrm{dark}}$ to be a naturalness measurement $S_d$. The lastly concerned naturalness-related features are defined as $\textit{F}_{\textit{n}}=\{\nu,\sigma^2,\textit{S}_\textit{d}\}$.

To summarize, on the basis of five respects of considerations which are composed of contrast, sharpness, brightness, colorfulness and naturalness of images, we elaborately extract a sum of 17 features. Towards readers' conveniences, all the features described above are listed in Table \ref{tab:01}.

\subsection{Quality Prediction}
\label{sec:2.2}
So far we have gained enhancement-related features, whose effectiveness will be discussed in Section \ref{sec:3}. These features however cannot offer a straightforward impression on how the quality of an enhanced image is. In this situation, a regression module converting 17 features into one quality score becomes desirable. The linear weighting combination is a simple and commonly used scheme. In order to integrate 17 features, at least 16 weights are required. Facing to such high-dimensional space of weighs, it is difficult to seek robust and reasonable values of parameters.

Another way to integrate features is to take advantage of dimensionality reduction tools, such as PCA and LLE [\ref{ref:50}]. But the extracted features play different roles in assessing the quality of enhanced images, and furthermore, they are also of different dimensions. This renders the use of dimensionality reduction a tough road.

Recently, a new strategy has been proposed towards finding the regression module in blind IQA designs [\ref{ref:51}]. To be more specific, in order to overcome the issue of overfitting, greater than 100,000 images are utilized as training samples to learn the regression module in our blind BIQME metric. Note that, in classical IQA researches, they usually report the median performance indices across 1,000 iterations of random 80\% train-20\% test procedure in a certain database [\ref{ref:14}, \ref{ref:17}, \ref{ref:16}, \ref{ref:18}] or they adopt the leave-one-out cross-validation methodology [\ref{ref:33}, \ref{ref:46}], for the purpose of verifying the effectiveness of their features. Of course we exploit the two manners above to verify the superiority of our enhancement-aware features as well in Section \ref{sec:3}. Nonetheless, due to limited visual scenes and only hundreds of images included in existing databases, these two manners readily cause overfitting in learning the regression module. So we deployed a valid strategy similar to that used in [\ref{ref:51a}]. We have first collected 1,642 images that contain 1242 natural scene images coming from Berkeley database [\ref{ref:52}] and high-quality subsets in PQD database [\ref{ref:53}] as well as 400 screen content images captured by ourselves with a screenshot tool\footnote{We will release the 400 screen content images online soon.}. These 1642 original images are absolutely content-independent of those in all the testing databases used in this research. Next we simulated enhanced images with eight typical global-based enhancement technologies akin to that employed in the CCID2014 database [\ref{ref:27}] and create 60 enhanced images for each original image. Including the 1642 original images, we eventually produce 100,162 images (much bigger than the size of the largest testing CCID2014 database that consists of 655 images) as training data.

How to label these generated images? In [\ref{ref:51}], Gu \emph{et al.} indicated that, rather than training on human opinion ratings, using predicted scores computed from high-performance FR-IQA methods as training labels is a good choice. The lately proposed PCQI metric was proven to highly correlate with subjective quality scores on enhancement-relevant databases [\ref{ref:32}], but it does not take the influence of colorfulness into consideration, which is obviously an important index of image quality. Based on this concern, we propose the Colorfulness-based PCQI (C-PCQI) metric:
\begin{equation}
\textrm{C-PCQI}=\frac{1}{M}\sum_{i=1}^MQ_{mi}(i)\cdot Q_{cc}(i)\cdot Q_{sd}(i)\cdot Q_{cs}(i)
\label{func:11a}
\end{equation}
where $Q_{mi}$, $Q_{cc}$ and $Q_{sd}$ respectively represent the similarity between the original and distorted images in terms of mean intensity, contrast change and structural distortion. More information about the definitions of these three terms can be found in [\ref{ref:32}]. $M$ is the number of pixels. $Q_{cs}$ measures the similarity of color saturation defined by
\begin{equation}
Q_{cs}(i)=\Big(\frac{2 ST_1\cdot ST_2+\zeta}{ST_1^2+ST_2^2+\zeta}\Big)^\varphi
\label{func:11b}
\end{equation}
where $ST_1$ and $ST_2$ stand for the color saturation of the original and distorted images, respectively. $\zeta$ is a very small constant number for avoiding division-by-zero and $\varphi$ is a fixed pooling index for stressing the areas which have remarkable changes of color saturation. We apply the C-PCQI scores of the 100,162 training images to replace human opinion ratings.

After the training set prepared, the famous support vector regression (SVR) is employed to learn the regression module in the proposed BIQME metric [\ref{ref:54}]. In general, traditional deep learning tools are not appropriate since there are only 17 features extracted. But it deserves to mention that a very good work has recently applied parallel computation of low-level features followed by a deep learning based regression [\ref{ref:54a}], and this strategy will be considered in our future work. Considering a training dataset $D=\{(x_1,y_1),...,(x_r,y_r)\}$, where $x_i$ and $y_i$, $i=1,...,r$, indicate a feature vector of $f_{01}$-$f_{17}$ in Table \ref{tab:01} and the target output of the $i$-th training image's C-PCQI score. Supposing parameters $t>0$ and $p>0$, we can express the standard form of SVR as
\begin{align}
\mathop{\textrm{minimize}}_{\mathbf{w},\delta,\mathbf{b},\mathbf{b'}}\quad\;\;& \frac{1}{2}||\mathbf{w}||_2^2+t\sum_{i=1}^r(b_i+b_i')\qquad\quad\,\,\,\\
\textrm{s.t.}\!\qquad\quad &\,\mathbf{w}^T\phi(x_i)+\delta-y_i\leq p+b_i, \nonumber\\
                             &\;y_i-\mathbf{w}^T\phi(x_i)-\delta\leq p+b_i',\nonumber\\
                             &\;b_i,b_i'\geq0,\,i=1,...,r.\nonumber
\label{func:12}
\end{align}
where $K(x_i,x_j)=\phi(x_i)^T\phi(x_j)$ is the kernel function, which is set to be the Radial Basis Function (RBF) kernel defined as $K(x_i,x_j)=\exp(-k\,||x_i-x_j||^2)$. Based on the training samples, our target is to determine the parameters $t$, $p$ and $k$ and thus find the associated regression module.

Finally, we also compare the proposed strategy with model distillation. The model distillation was a recently proposed concept in deep learning. Once the cumbersome model has been trained, a different kind of training called ``distillation'' can be used to transfer the knowledge from the cumbersome model to a small model that is more suitable for deployment [\ref{ref:54b}]. Compared with model distillation, the proposed strategy is close to a data-fitting adaption. That is, we deploy a high-performance FR-IQA model, which can approximate ``ground truth'', to learn the features to derive a NR-IQA model based on big-data training samples.

\section{Experimental Results and Discussions}
\label{sec:3}
In this section we will pay our attention to evaluating and comparing the performance of the proposed blind BIQME metric with up to 16 state-of-the-art IQA approaches on nine enhancement-related databases.

\subsection{Experimental Setup}
\label{sec:3.1}
Quality Measures. Recent years have seen an enumerous number of IQA measures, most of which not only obtain high performance accuracy but only consume few implementation time. In this research, we choose the following four types of methods. The first type includes FSIM [\ref{ref:06}], LTG [\ref{ref:08}], VSI [\ref{ref:09}], and PSIM [\ref{ref:09b}], which all belong to FR metrics and acquire superior performance on popular databases. The second type consists of two RR-IQA models, RRED [\ref{ref:12}] and FTQM [\ref{ref:13}]. The third type contains BRISQUE [\ref{ref:14}], NFERM [\ref{ref:17}], NIQE [\ref{ref:15}], IL-NIQE [\ref{ref:19}] and BQMS [\ref{ref:51}] without access to original references in assessing the visual quality of images. The last one consists of FR C-PCQI, RR RIQMC [\ref{ref:27}], RR QMC [\ref{ref:31}], blind FANG [\ref{ref:33}], and blind GISTCM [\ref{ref:34}], which are dedicated to enhanced IQA tasks.

Testing Datasets. To the best of our knowledge, there exist nine main relevant subjective image databases. The first two are CID2013 and CCID2014 databases [\ref{ref:30}, \ref{ref:27}], which have been constructed particularly for image quality evaluation of contrast change in Shanghai Jiao Tong University during the years 2013-2014. The two databases encompass 400 and 655 images through six and eight contrast alteration technologies, respectively. The second group is composed of four contrast enhancement-related subsets in TID2008, CSIQ, TID2013 and SIQAD databases [\ref{ref:24}, \ref{ref:25}, \ref{ref:26}, \ref{ref:55}]. There are 200, 116, 250 and 140 images in the aforementioned four subsets. The last three subsets are completed by Peking University in the year of 2013 [\ref{ref:34}]. Each of the three subsets includes 500 images, separately generated by enhancing haze, underwater and low-light images. Interested readers can be directed to [\ref{ref:24}, \ref{ref:25}, \ref{ref:26}, \ref{ref:27}, \ref{ref:30}, \ref{ref:34}, \ref{ref:55}] for detailed information of the nine datasets used in our work.

\begin{figure*}[!t]
\vspace{-0.08cm}
\small
\centering
\includegraphics[width=5.5cm]{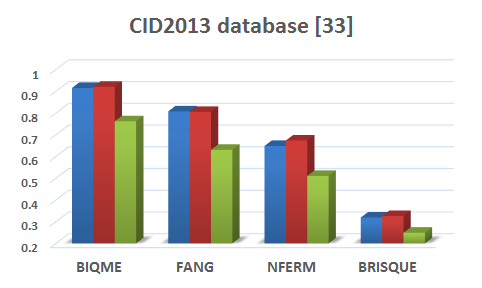}\hspace{0.1cm}
\includegraphics[width=5.5cm]{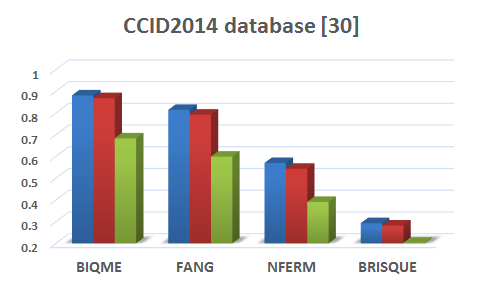}\hspace{0.1cm}
\includegraphics[width=5.5cm]{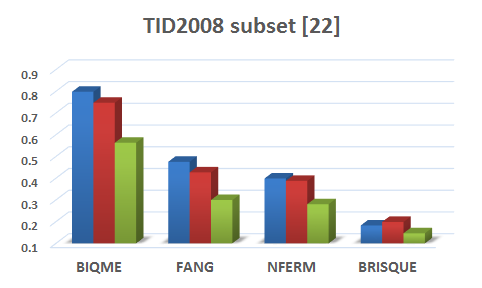}\\
\vspace{0.075cm}
\includegraphics[width=5.5cm]{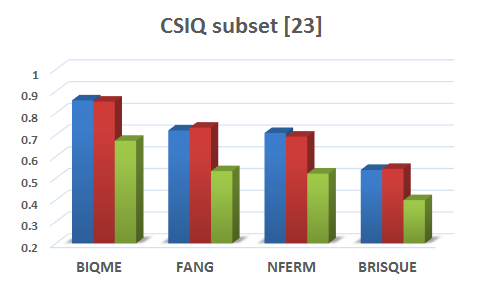}\hspace{0.1cm}
\includegraphics[width=5.5cm]{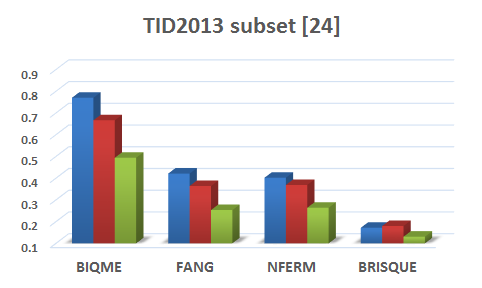}\hspace{0.1cm}
\includegraphics[width=5.5cm]{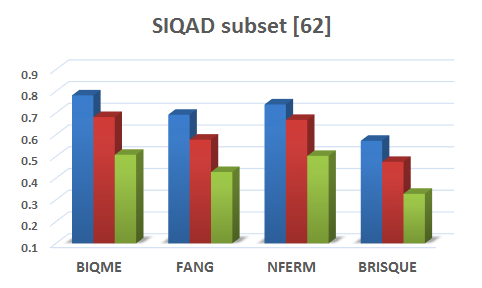}\\
\vspace{-0.075cm}
\caption{\small Performance of BIQME (proposed), FANG [\ref{ref:33}], NFERM [\ref{ref:17}] and BRISQUE [\ref{ref:14}] metrics on CID2013, CCID2014, TID2008, CSIQ, TID2013 and SIQAD datasets. Blue, red and green bars respectively represent PLC, SRC and KRC indices.}
\label{fig:03}
\vspace{0.2cm}
\end{figure*}

\begin{figure*}[!t]
\vspace{0.1cm}
\small
\centering
\includegraphics[width=3.10cm]{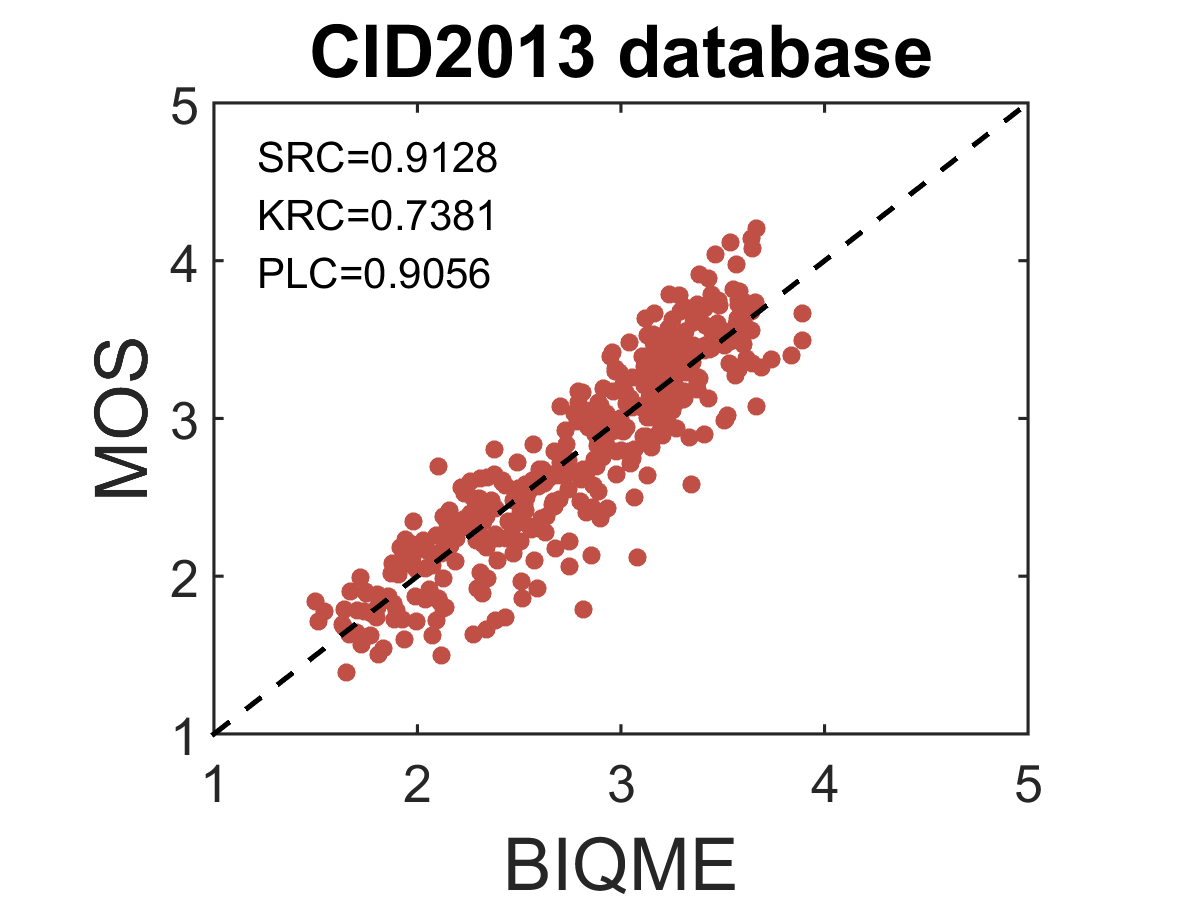}\hspace{-0.350cm}
\includegraphics[width=3.10cm]{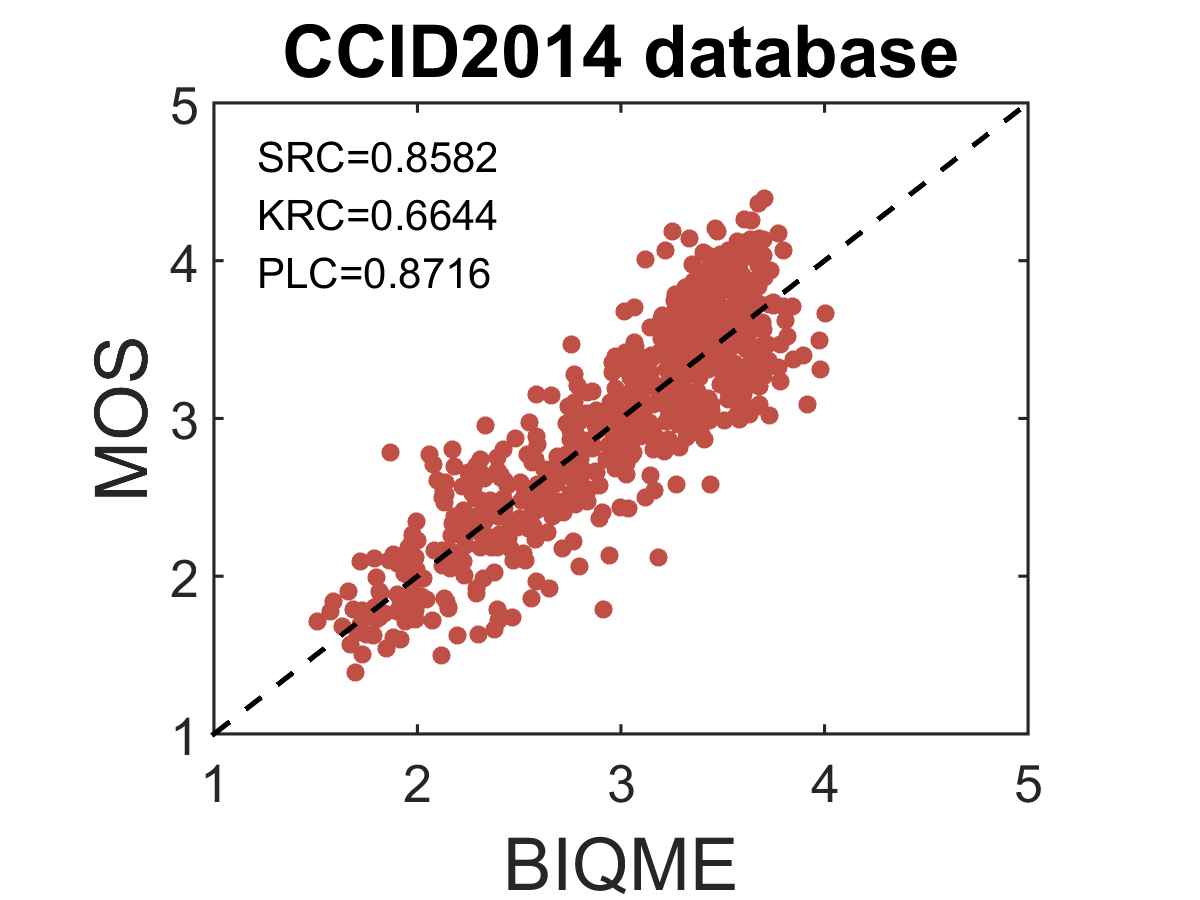}\hspace{-0.350cm}
\includegraphics[width=3.10cm]{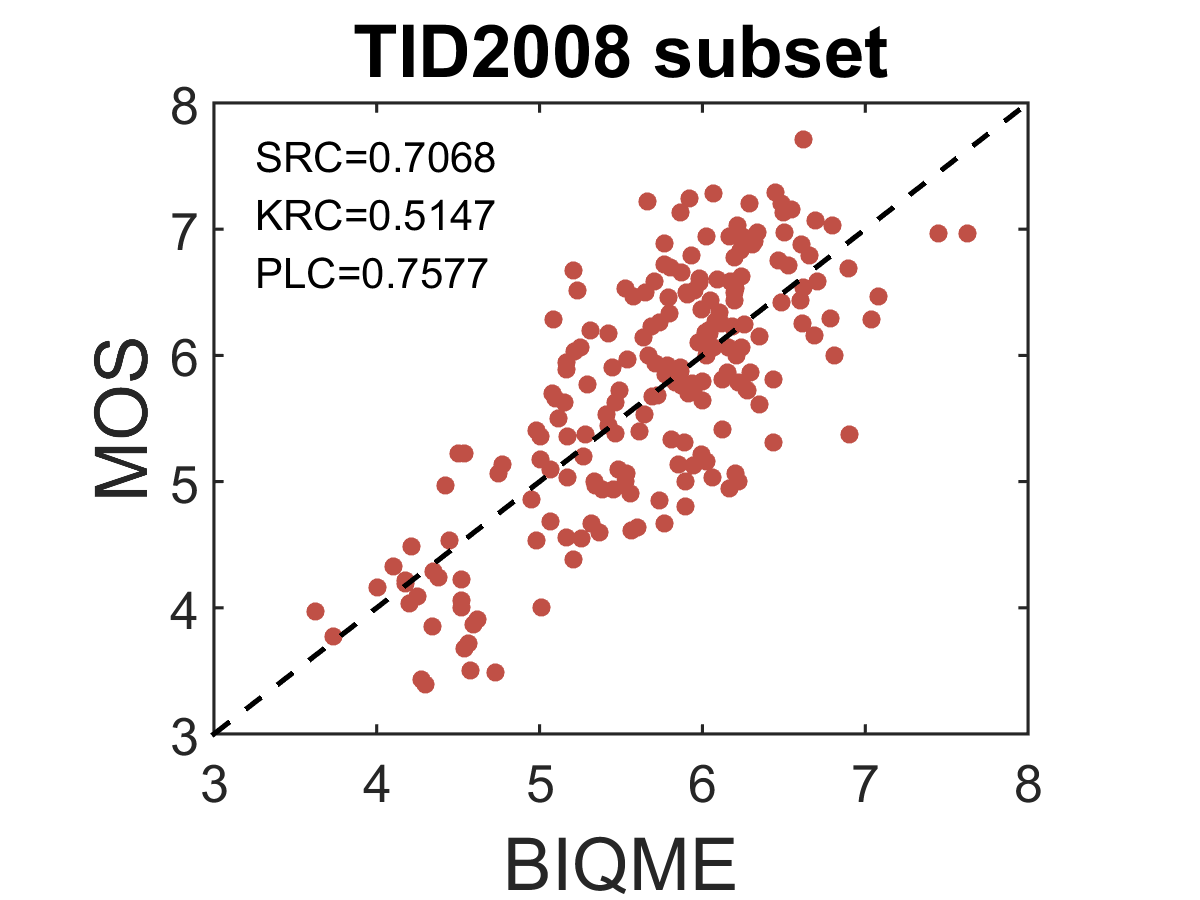}\hspace{-0.350cm}
\includegraphics[width=3.10cm]{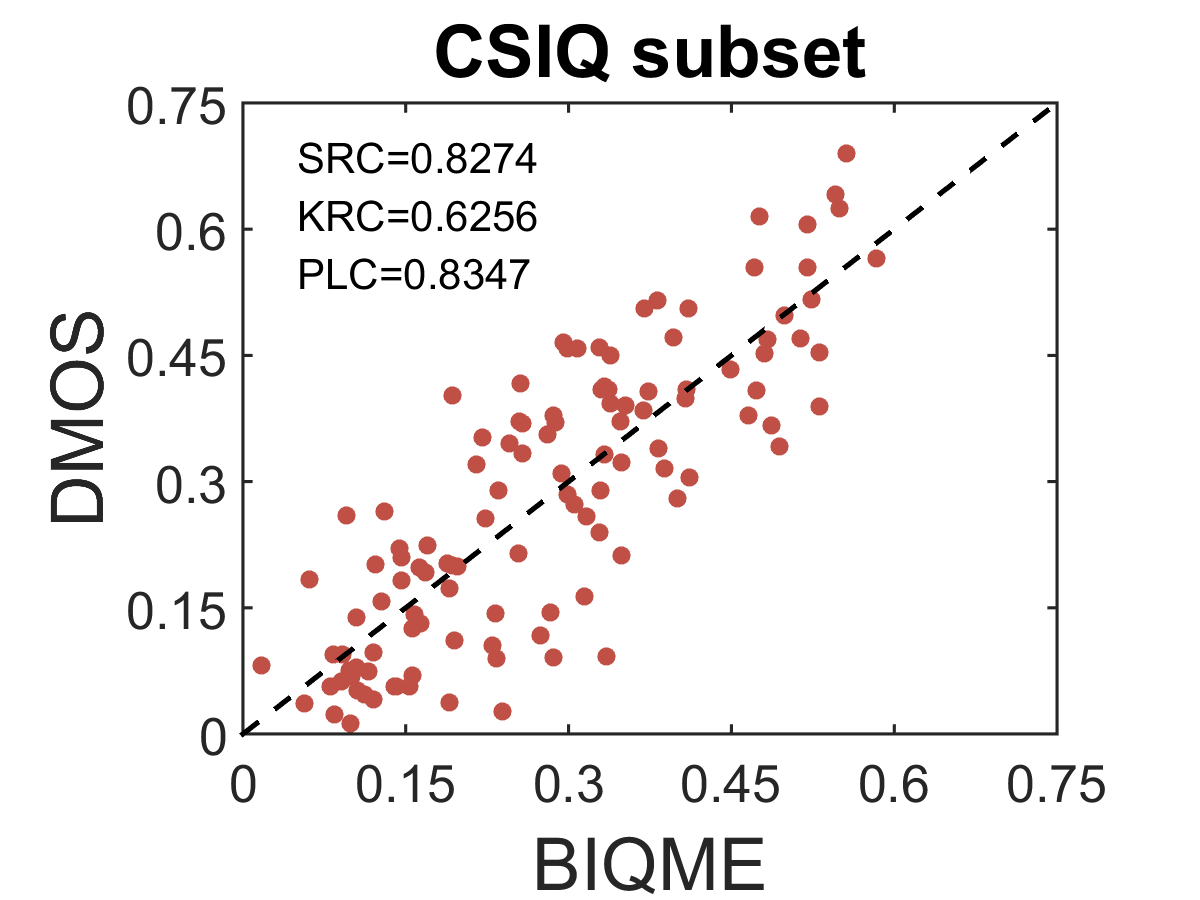}\hspace{-0.350cm}
\includegraphics[width=3.10cm]{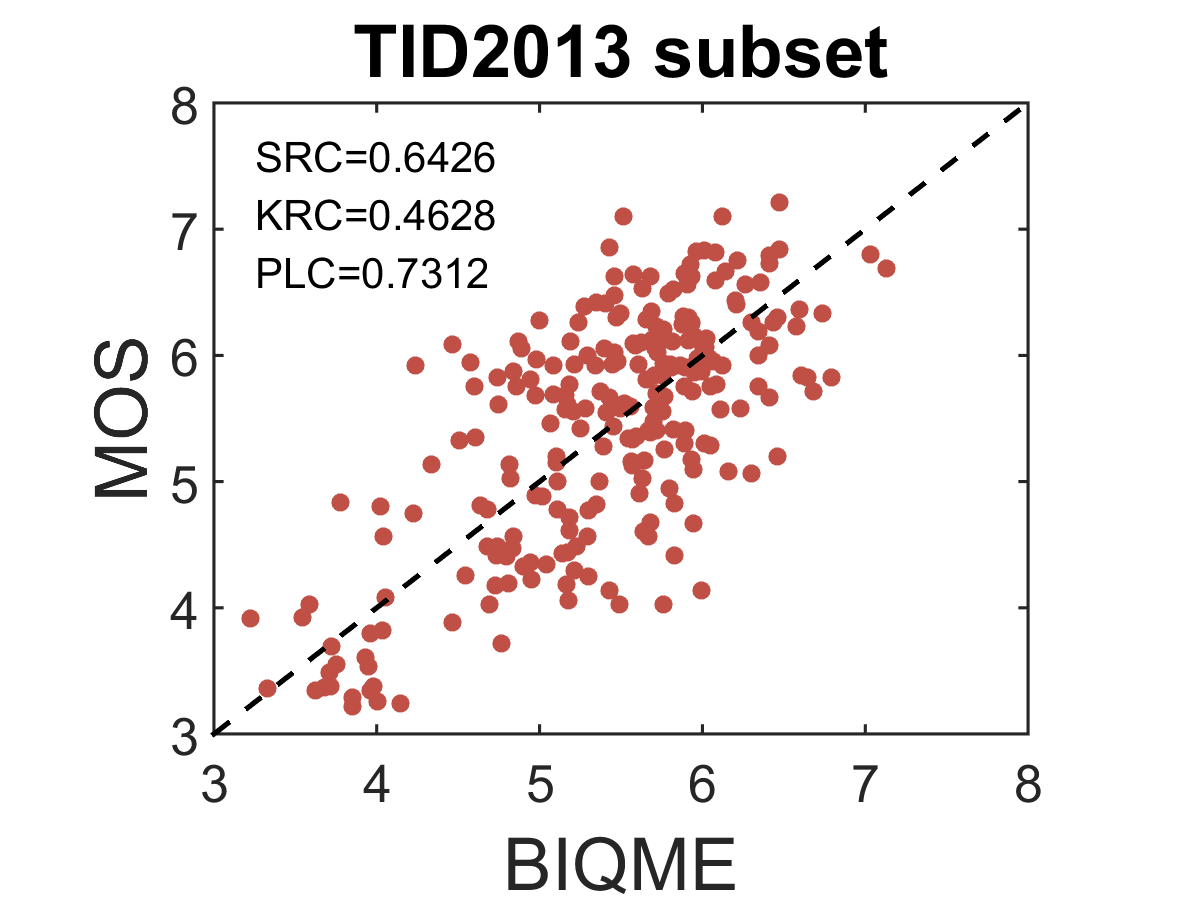}\hspace{-0.350cm}
\includegraphics[width=3.10cm]{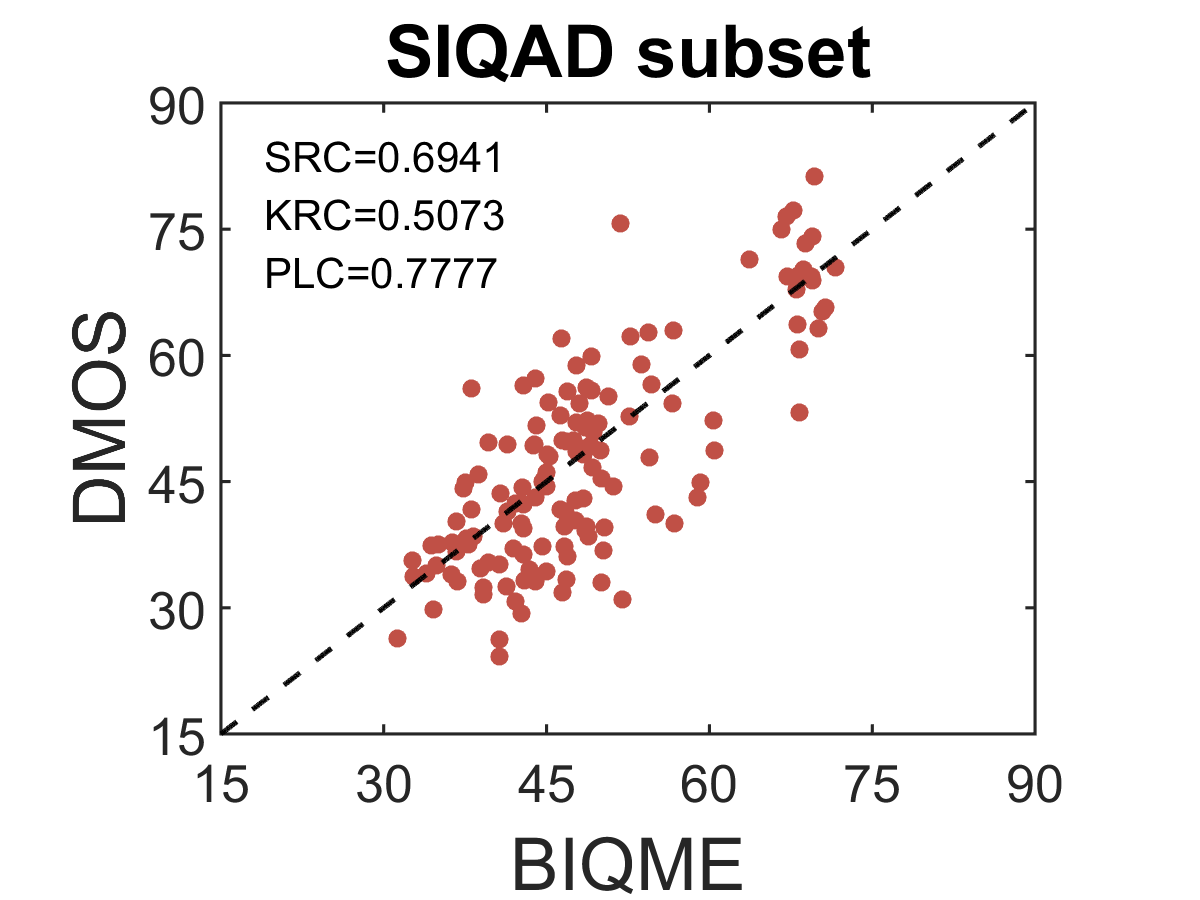}\\
\vspace{0.175cm}
\includegraphics[width=3.10cm]{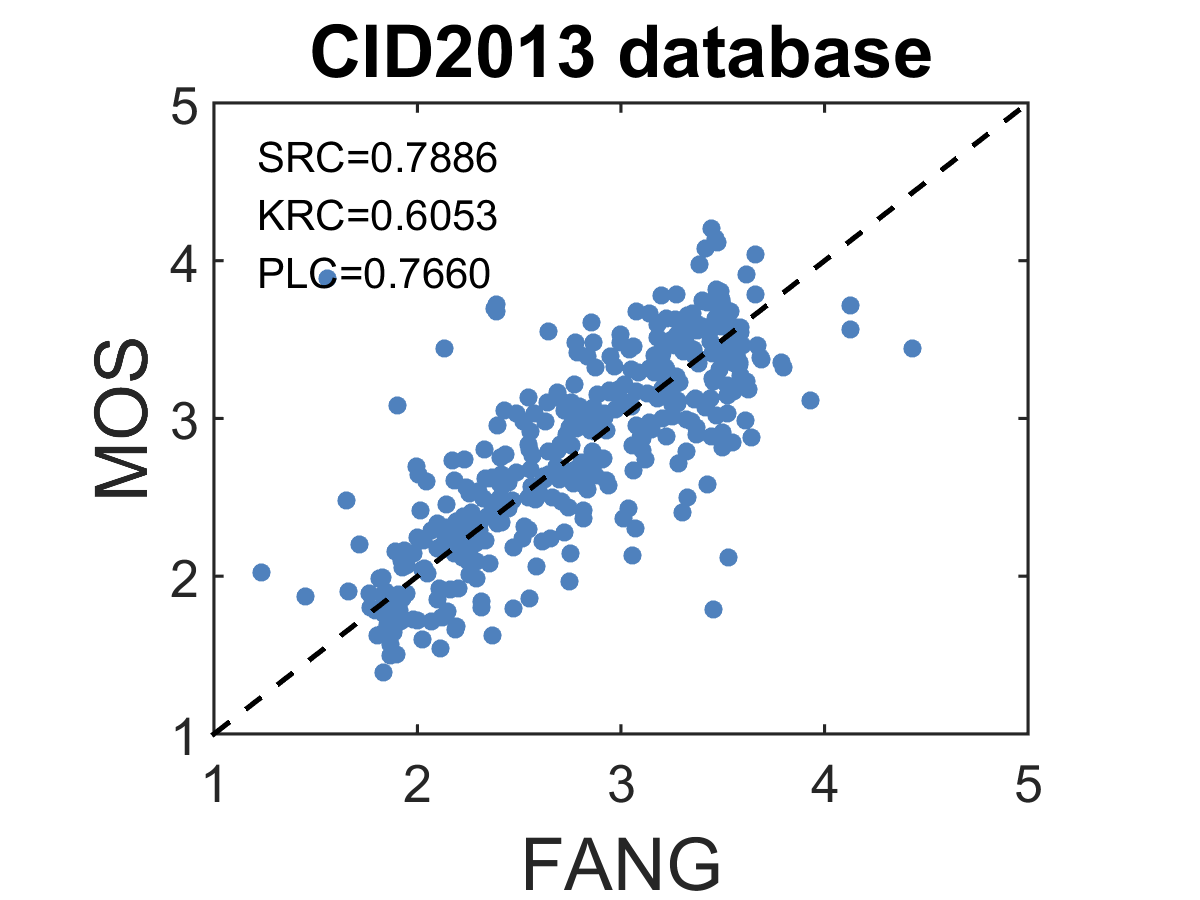}\hspace{-0.350cm}
\includegraphics[width=3.10cm]{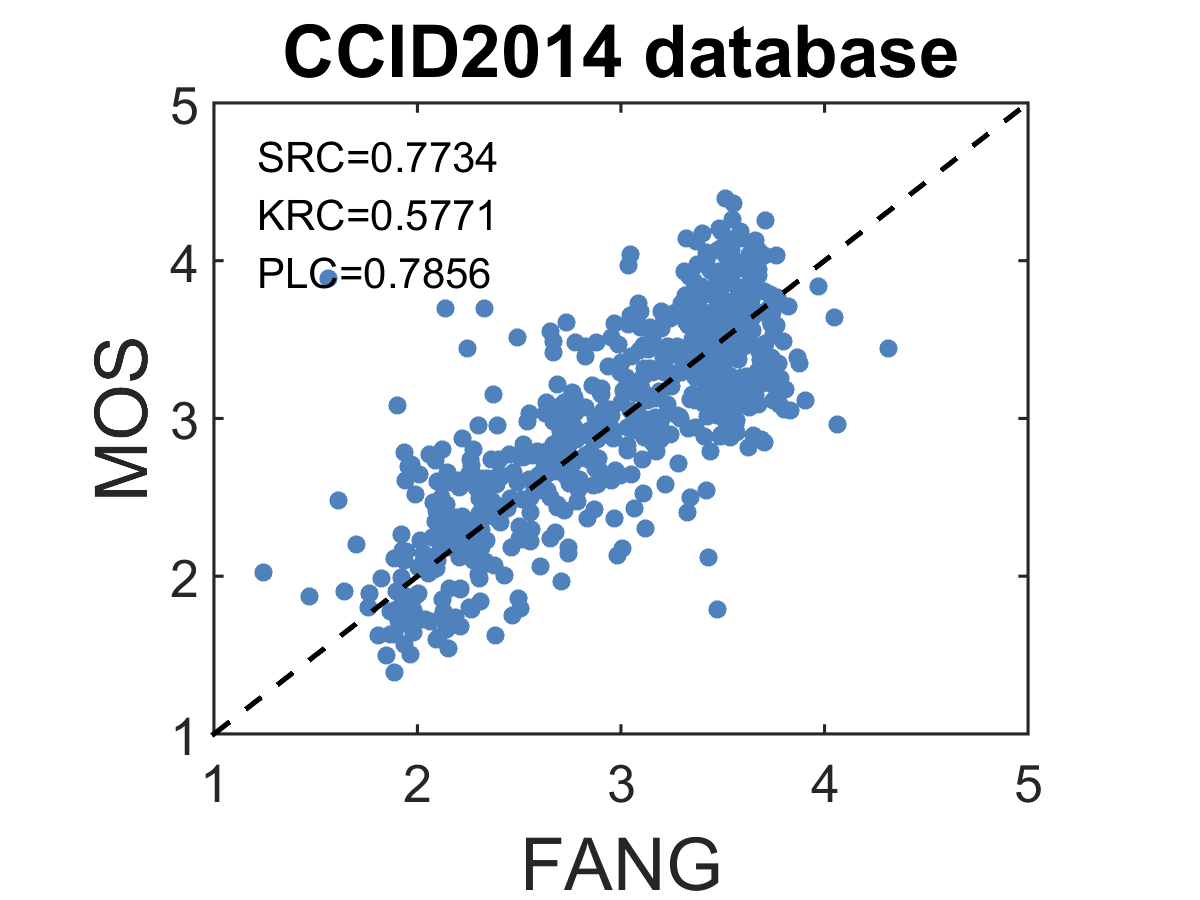}\hspace{-0.350cm}
\includegraphics[width=3.10cm]{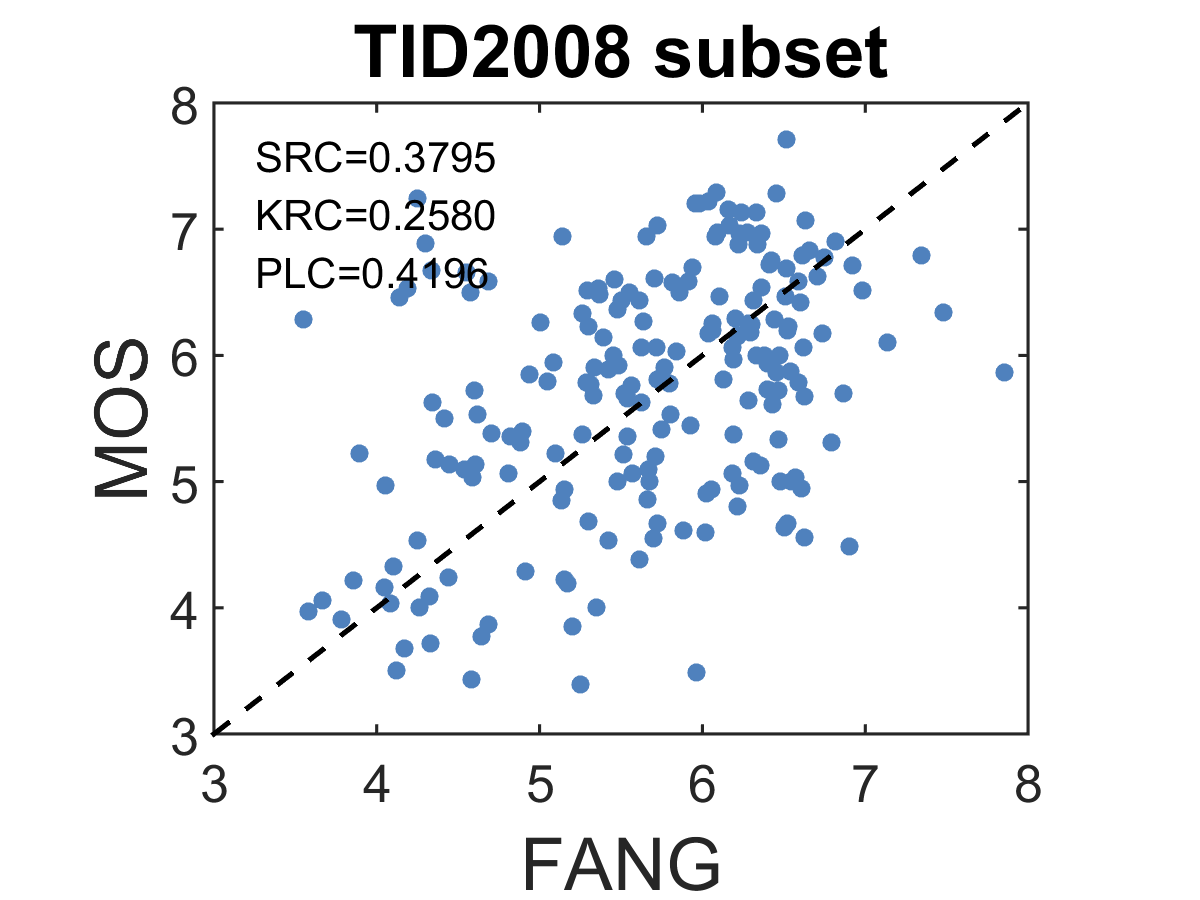}\hspace{-0.350cm}
\includegraphics[width=3.10cm]{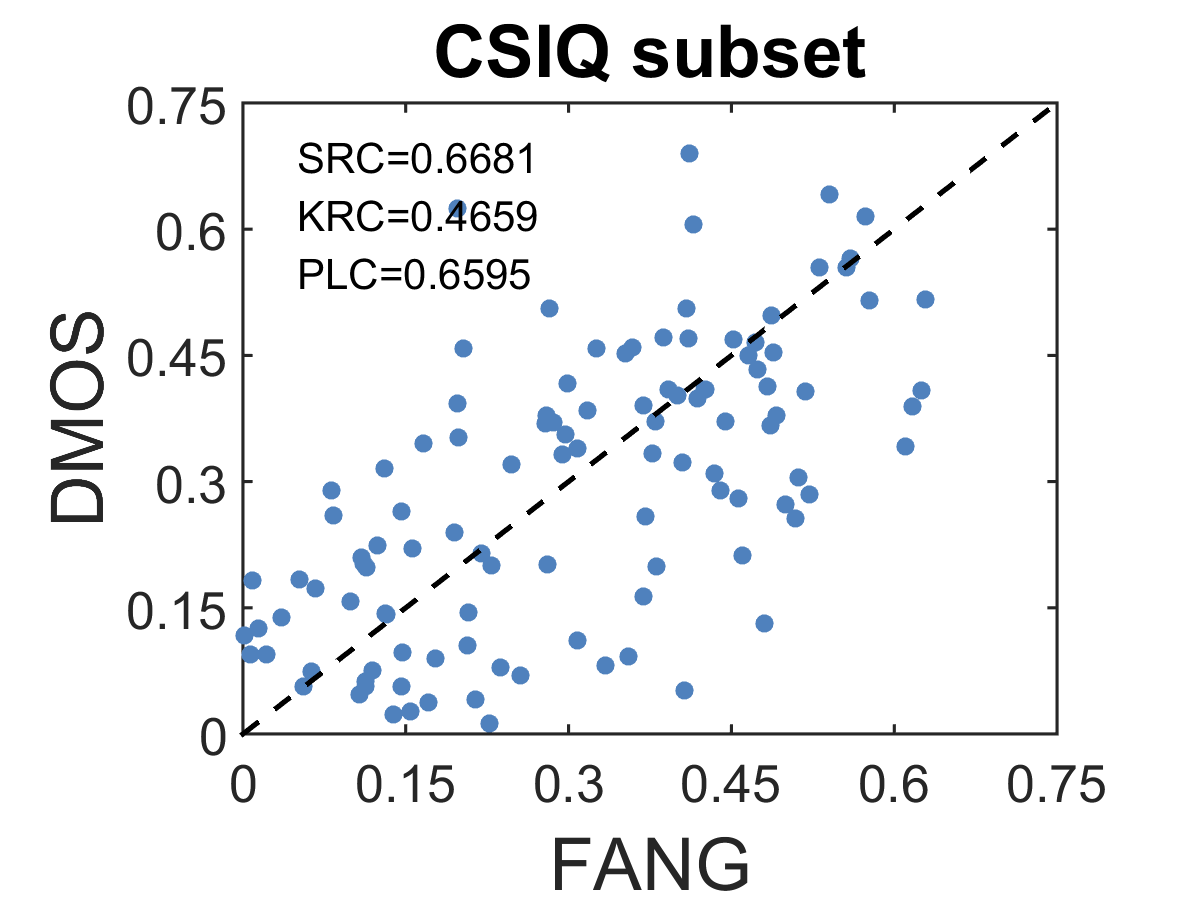}\hspace{-0.350cm}
\includegraphics[width=3.10cm]{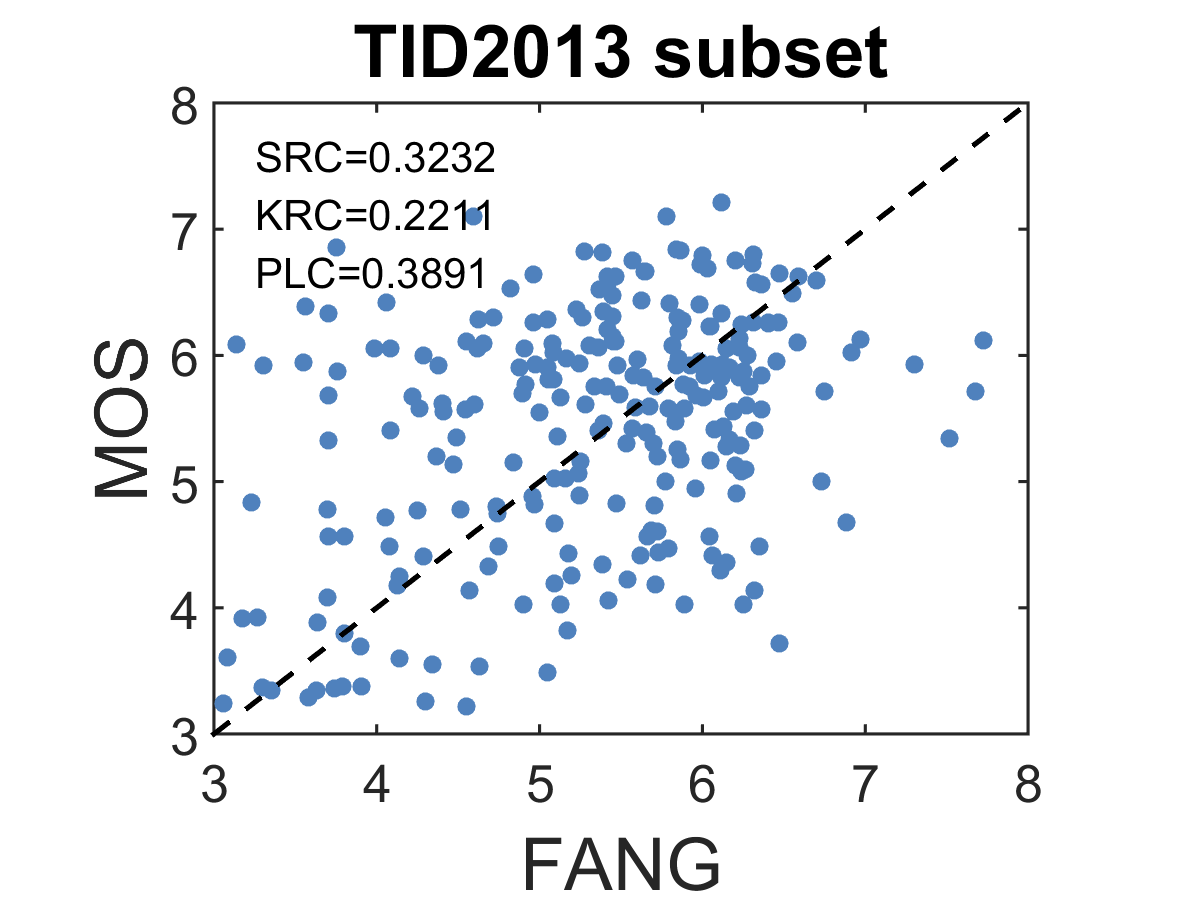}\hspace{-0.350cm}
\includegraphics[width=3.10cm]{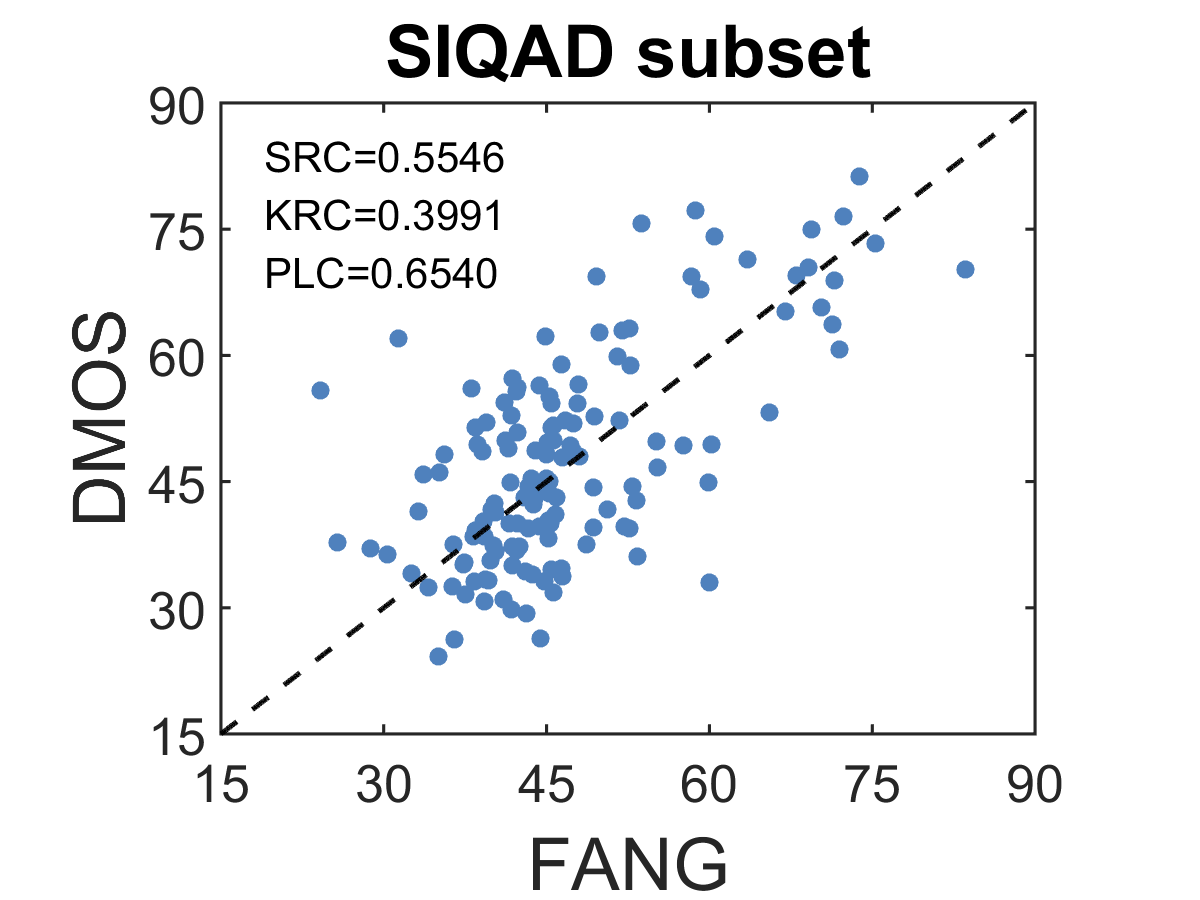}\\
\vspace{-0.175cm}
\caption{\small Scatter plots of BIQME (proposed) and FANG [\ref{ref:33}] using a leave-one-out cross-validation experiment on six datasets.}
\label{fig:04}
\vspace{0.05cm}
\end{figure*}

Performance Benchmarking. In general, there are three representative evaluation metrics for correlation performance measure and comparison in most IQA studies. The first one is Spearman rank order correlation coefficient (SRC) or rank correlation coefficient, which is a non-parametric test\footnote{Non-parametric indicates a test does not rely on any assumption on the distributions of two variables.} towards calculating the degree of association between two variables from the angle of prediction monotonicity. The second one is another non-parametric monotonicity index, Kendall's rank-order correlation coefficient (KRC), focusing on evaluating the strength of dependence of two variables. Compared with SRC, KRC has stricter demands, for example, both testing variables must be ordinal. The third criterion is Pearson linear correlation coefficient (PLC), which is commonly abbreviated to linear correlation coefficient. PLC estimates the prediction accuracy between two variables. It requires to stress that the nonlinearity of objective quality scores should be eliminated using regression functions before computing PLC index. Two typical regression functions are the four-parameter function
\begin{equation}
\mathbf{g}(\mathbf{q})=\frac{\tau_1-\tau_2}{1+\exp{(-\frac{\mathbf{q}-\tau_3}{\tau_4})}}+\tau_2
\label{func:13}
\end{equation}
and the five-parameter function
\begin{equation}
\mathbf{g}(\mathbf{q})=\tau_1\bigg(0.5-\frac{1}{1+\exp[\tau_2(\mathbf{q}-\tau_3)]}\bigg)+\tau_4\mathbf{q}+\tau_5
\label{func:14}
\end{equation}
where $\mathbf{q}$ and $\mathbf{g}(\mathbf{q})$ are the vectors of raw objective quality scores and converted scores after the nonlinear regression of (\ref{func:13}) or (\ref{func:14}); we use the curve fitting process to compute the values of model parameters $\{\tau_1,...,\tau_4\}$ or $\{\tau_1,...,\tau_5\}$. This paper adopts the five-parameter logistic function. Of the three performance evaluation criteria, a value approaching to one for PLC, SRC and KRC means the superior performance in line with human opinion ratings.

\begin{table}[!t]
\vspace{-0.08cm}
\center
\caption{\small Comparison on haze, underwater and low-light subsets.}
\label{tab:02}
\vspace{0.2cm}
\center
\begin{tabular}{|c|c|c|c|c|}
\hline
SRC&Length&\;\;\;\;Haze\;\;\;\;&Under water&Low light\\
\hline
BIQME           (Pro.) &  17 & 0.7290 & 0.8171 & 0.9123 \\
BRISQUE [\ref{ref:14}] &  36 & 0.4179 & 0.4781 & 0.4461 \\
NFERM   [\ref{ref:17}] &  23 & 0.4988 & 0.6334 & 0.7925 \\
FANG    [\ref{ref:33}] &   5 & 0.5196 & 0.1467 & 0.8316 \\
GISTCM  [\ref{ref:34}] & 521 & 0.6302 & 0.7858 & 0.9155 \\
\hline
\end{tabular}
\end{table}

\begin{table*}[!t]
\vspace{0.03cm}
\center
\caption{\small Performance comparison of 14 state-of-the-art IQA measures. We highlight the top metric in each type.}
\label{tab:03}
\vspace{0.2cm}
\center
\begin{tabular}{|l|c|ccc|ccc|ccc|ccc|}
\hline
\multirow{2}{*}{Models\!}&\multirow{2}{*}{\!\!Type\!\!}&\multicolumn{3}{c|}{CID2013 [\ref{ref:30}]}&\multicolumn{3}{c|}{CCID2014 [\ref{ref:27}]}&\multicolumn{3}{c|}{TID2008 [\ref{ref:24}]}&\multicolumn{3}{c|}{CSIQ [\ref{ref:25}]}\\
\cline{3-14}&&PLC&\!\!\!SRC\!\!\!&KRC&PLC&\!\!\!SRC\!\!\!&KRC&PLC&\!\!\!SRC\!\!\!&KRC&PLC&\!\!\!SRC\!\!\!&KRC\\
\hline
FSIM\!&FR&0.8574&\!\!\!0.8486\!\!\!&0.6663&0.8201&\!\!\!0.7658\!\!\!&0.5707&0.6880&\!\!\!0.4403\!\!\!&0.3348&0.9378&\!\!\!0.9420\!\!\!&0.7883\\
LTG\!&FR&0.8656&\!\!\!0.8605\!\!\!&0.6723&0.8384&\!\!\!0.7901\!\!\!&0.5938&0.6795&\!\!\!0.4655\!\!\!&0.3285&\textbf{0.9560}&\!\!\!0.9414\!\!\!&0.7880\\
VSI\!&FR&0.8571&\!\!\!0.8506\!\!\!&0.6579&0.8209&\!\!\!0.7734\!\!\!&0.5736&0.6819&\!\!\!0.4571\!\!\!&0.3450&0.9532&\!\!\!\textbf{0.9504}\!\!\!&\textbf{0.8096}\\
PSIM\!&FR&0.8604&\!\!\!0.8541\!\!\!&0.6666&0.8386&\!\!\!0.8004\!\!\!&0.6038&0.6106&\!\!\!0.4573\!\!\!&0.3202&0.9447&\!\!\!0.9336\!\!\!&0.7718\\
C-PCQI\!\!\!\!&FR&\textbf{0.9247}&\!\!\!\textbf{0.9260}\!\!\!&\textbf{0.7586}&\textbf{0.8885}&\!\!\!\textbf{0.8754}\!\!\!&\textbf{0.6858}&\textbf{0.9061}&\!\!\!\textbf{0.8782}\!\!\!&\textbf{0.7016}&0.9454&\!\!\!0.9394\!\!\!&0.7820\\
\hline
RRED\!&RR&0.7295&\!\!\!0.7218\!\!\!&0.5254&0.7064&\!\!\!0.6595\!\!\!&0.4677&0.5278&\!\!\!0.2320\!\!\!&0.1693&0.9415&\!\!\!0.9382\!\!\!&0.7838\\
FTQM\!&RR&0.8164&\!\!\!0.8047\!\!\!&0.6125&0.7885&\!\!\!0.7292\!\!\!&0.5330&0.6845&\!\!\!0.3006\!\!\!&0.1854&0.9552&\!\!\!0.9532\!\!\!&0.8129\\
RIQMC\!&RR&0.8995&\!\!\!0.9005\!\!\!&0.7162&0.8726&\!\!\!0.8465\!\!\!&0.6507&\textbf{0.8585}&\!\!\!\textbf{0.8095}\!\!\!&\textbf{0.6224}&\textbf{0.9652}&\!\!\!\textbf{0.9579}\!\!\!&\textbf{0.8279}\\
QMC\!&RR&\textbf{0.9309}&\!\!\!\textbf{0.9340}\!\!\!&\textbf{0.7713}&\textbf{0.8960}&\!\!\!\textbf{0.8722}\!\!\!&\textbf{0.6872}&0.7688&\!\!\!0.7340\!\!\!&0.5520&0.9622&\!\!\!0.9554\!\!\!&0.8207\\
\hline
NIQE\!&NR&0.4648&\!\!\!0.3929\!\!\!&0.2709&0.4694&\!\!\!0.3655\!\!\!&0.2494&0.0979&\!\!\!0.0223\!\!\!&0.0187&0.3019&\!\!\!0.2444\!\!\!&0.1613\\
IL-NIQE\!&NR&0.5682&\!\!\!0.5273\!\!\!&0.3708&0.5764&\!\!\!0.5121\!\!\!&0.3590&0.2244&\!\!\!0.1833\!\!\!&0.1223&0.5468&\!\!\!0.5005\!\!\!&0.3510\\
BQMS\!&NR&0.5733&\!\!\!0.4624\!\!\!&0.3196&0.5742&\!\!\!0.4381\!\!\!&0.3039&0.2450&\!\!\!0.1539\!\!\!&0.1024&0.3259&\!\!\!0.3178\!\!\!&0.2241\\
FANG\!&NR&0.7904&\!\!\!0.8006\!\!\!&0.5893&0.7890&\!\!\!0.7822\!\!\!&0.5684&0.2737&\!\!\!0.2666\!\!\!&0.1785&0.1762&\!\!\!0.1870\!\!\!&0.1175\\
BIQME\!&NR&\textbf{0.9004}&\!\!\!\textbf{0.9023}\!\!\!&\textbf{0.7223}&\textbf{0.8588}&\!\!\!\textbf{0.8309}\!\!\!&\textbf{0.6305}&\textbf{0.7476}&\!\!\!\textbf{0.6980}\!\!\!&\textbf{0.5123}&\textbf{0.8129}&\!\!\!\textbf{0.7851}\!\!\!&\textbf{0.5980}\\
\hline\noalign{\smallskip\smallskip}
\hline
\multirow{2}{*}{Models\!}&\multirow{2}{*}{\!\!Type\!\!}&\multicolumn{3}{c|}{TID2013 [\ref{ref:26}]}&\multicolumn{3}{c|}{SIQAD [\ref{ref:55}]}&\multicolumn{3}{c|}{Direct mean}&\multicolumn{3}{c|}{Weighted mean}\\
\cline{3-14}&&PLC&\!\!\!SRC\!\!\!&KRC&PLC&\!\!\!SRC\!\!\!&KRC&PLC&\!\!\!SRC\!\!\!&KRC&PLC&\!\!\!SRC\!\!\!&KRC\\
\hline
FSIM\!&FR&0.6819&\!\!\!0.4413\!\!\!&0.3588&\textbf{0.8222}&\!\!\!0.7150\!\!\!&0.5328&0.8012&\!\!\!0.6921\!\!\!&0.5419&0.8019&\!\!\!0.7091\!\!\!&0.5468\\
LTG\!&FR&0.6749&\!\!\!0.4639\!\!\!&0.3458&0.7820&\!\!\!0.6539\!\!\!&0.4773&0.7994&\!\!\!0.6959\!\!\!&0.5343&0.8066&\!\!\!0.7221\!\!\!&0.5498\\
VSI\!&FR&0.6785&\!\!\!0.4643\!\!\!&0.3705&0.7734&\!\!\!0.6461\!\!\!&0.4728&0.7942&\!\!\!0.6903\!\!\!&0.5382&0.7981&\!\!\!0.7127\!\!\!&0.5455\\
PSIM\!&FR&0.6092&\!\!\!0.4542\!\!\!&0.3347&0.7098&\!\!\!0.5864\!\!\!&0.4146&0.7622&\!\!\!0.6810\!\!\!&0.5186&0.7819&\!\!\!0.7162\!\!\!&0.5437\\
C-PCQI\!\!\!\!&FR&\textbf{0.9175}&\!\!\!\textbf{0.8805}\!\!\!&\textbf{0.7074}&0.8127&\!\!\!\textbf{0.7447}\!\!\!&\textbf{0.5624}&\textbf{0.8991}&\!\!\!\textbf{0.8740}\!\!\!&\textbf{0.6996}&\textbf{0.9006}&\!\!\!\textbf{0.8817}\!\!\!&\textbf{0.7037}\\
\hline
RRED\!&RR&0.5606&\!\!\!0.3068\!\!\!&0.2419&0.7347&\!\!\!0.5601\!\!\!&0.3942&0.7001&\!\!\!0.5697\!\!\!&0.4304&0.6884&\!\!\!0.5855\!\!\!&0.4299\\
FTQM\!&RR&0.7697&\!\!\!0.6095\!\!\!&0.4685&\textbf{0.8216}&\!\!\!\textbf{0.6976}\!\!\!&\textbf{0.5205}&0.8060&\!\!\!0.6825\!\!\!&0.5221&0.7940&\!\!\!0.6929\!\!\!&0.5199\\
RIQMC\!&RR&\textbf{0.8651}&\!\!\!\textbf{0.8044}\!\!\!&\textbf{0.6178}&0.5479&\!\!\!0.4506\!\!\!&0.3139&\textbf{0.8348}&\!\!\!\textbf{0.7949}\!\!\!&\textbf{0.6248}&\textbf{0.8563}&\!\!\!\textbf{0.8244}\!\!\!&\textbf{0.6426}\\
QMC\!&RR&0.7713&\!\!\!0.7153\!\!\!&0.5364&0.2610&\!\!\!0.2485\!\!\!&0.1653&0.7650&\!\!\!0.7432\!\!\!&0.5888&0.8256&\!\!\!0.8042\!\!\!&0.6369\\
\hline
NIQE\!&NR&0.0985&\!\!\!0.0788\!\!\!&0.0522&0.1364&\!\!\!0.1607\!\!\!&0.1137&0.2615&\!\!\!0.2108\!\!\!&0.1444&0.3360&\!\!\!0.2678\!\!\!&0.1835\\
IL-NIQE\!&NR&0.2275&\!\!\!0.1517\!\!\!&0.1030&0.3044&\!\!\!0.2491\!\!\!&0.1786&0.4080&\!\!\!0.3540\!\!\!&0.2475&0.4615&\!\!\!0.4054\!\!\!&0.2836\\
BQMS\!&NR&0.2514&\!\!\!0.1885\!\!\!&0.1259&0.3146&\!\!\!0.2450\!\!\!&0.1642&0.3807&\!\!\!0.3010\!\!\!&0.2067&0.4538&\!\!\!0.3526\!\!\!&0.2429\\
FANG\!&NR&0.2941&\!\!\!0.2675\!\!\!&0.1742&0.2768&\!\!\!0.1904\!\!\!&0.1324&0.4334&\!\!\!0.4157\!\!\!&0.2934&0.5794&\!\!\!0.5685\!\!\!&0.4085\\
BIQME\!&NR&\textbf{0.7259}&\!\!\!\textbf{0.6444}\!\!\!&\textbf{0.4693}&\textbf{0.7860}&\!\!\!\textbf{0.6783}\!\!\!&\textbf{0.4954}&\textbf{0.8053}&\!\!\!\textbf{0.7565}\!\!\!&\textbf{0.5713}&\textbf{0.8279}&\!\!\!\textbf{0.7904}\!\!\!&\textbf{0.6022}\\
\hline
\end{tabular}
\end{table*}

\begin{table*}[!t]
\vspace{0.2cm}
\center
\caption{\small Mean implementation time on all the 665 images in the CCID2014 database.}
\label{tab:04}
\vspace{0.2cm}
\center
\begin{tabular}{|c|c|c|c|c|c|c|c|}
\hline\noalign{}
IQA models&FSIM&\;\;LTG\;\;&\;\;VSI\;\;&\;\;PSIM\;\;&C-PCQI&RRED&FTQM\\
\hline
Time (second/image)&0.675&0.045&0.294&0.065&0.373&1.536&0.592\\
\hline
\noalign{\smallskip\smallskip}
\hline
IQA models&RIQMC&QMC&NIQE&IL-NIQE&BQMS&FANG&BIQME\\
\hline
Time (second/image)&0.867&0.010&0.450&3.064&90.72&0.693&0.906\\
\hline\noalign{}
\end{tabular}
\end{table*}

\subsection{Performance Results}
\label{sec:3.2}
Effectiveness of Features. We deploy two significant tests to measure the effectiveness of features. Firstly, inspired by [\ref{ref:14}, \ref{ref:17}, \ref{ref:16}], each testing dataset was randomly separated into two teams based on image scenes. We take the TID2008 subset as an example. Team 1 contains 160 training images corresponding to 20 original images and Team 2 contains 40 testing images corresponding to the remaining 5 original images. Using the 17 extracted features, the regression module is trained on the 80\% data from Team 1 and is employed to conduct performance evaluations on the 20\% data from Team 2. This procedure of random 80\% train-20\% test is repeated 1,000 times before median performance measures across the 1,000 iterations are provided for comparison. We respectively apply the aforesaid test on the former six datasets and list the results in Fig. \ref{fig:03}. Three representative NR-IQA measures, including BRISQUE, NFERM and FANG methods, satisfy the requirement of this experiment, so we also include them and report their results in Fig. \ref{fig:03}. On the last three subsets about dehaze images, enhanced underwater images and enhanced low-light images, we perform the same experiment with that used in [\ref{ref:34}]. SRC results are given in Table \ref{tab:02}. One can see that the proposed BIQME metric with a few features has attained encouraging performance, especially for contrast-changed images and enhanced haze images.

The second test exploits a leave-one-out cross-validation, akin to [\ref{ref:46}], for evaluating and comparing the effectiveness of features. More concretely, we also take the TID2008 subset to briefly illustrate how to carry out the leave-one-out cross-validation experiment. As for 8 testing images associated to one particular original image, we learn the regression module with other 192 training image associated to the rest 24 original images followed by predicting quality scores of the 8 image above. Likewise, we can obtain the quality measures of all 200 images on the TID2008 subset. Following this, the quality scores of objective IQA models on the entire images in other datasets can be yielded. This paper just compares our BIQME algorithm and the recently devised FANG metric dedicated to IQA of contrast adjustment because in most conditions they outperform the others. We just choose CID2013, CCID2014, TID2008, CSIQ, TID2013 and SIQAD datasets that meet the requirement of conducting the leave-one-out cross-validation. Results of experiments are illustrated in Fig. \ref{fig:04} in the manner of scatter plots. Towards convenient comparisons, we further label the numerical results on each scatter plot. As seen, both as blind IQA metrics, the proposed BIQME generates more reliable quality predictions, i.e. the sample points are closer to the black diagonal lines (indicating perfect performance), constantly and largely superior to the FANG.

\begin{figure*}[!t]
\center
\small
\vspace{-0.15cm}
\includegraphics[height=0.19\linewidth]{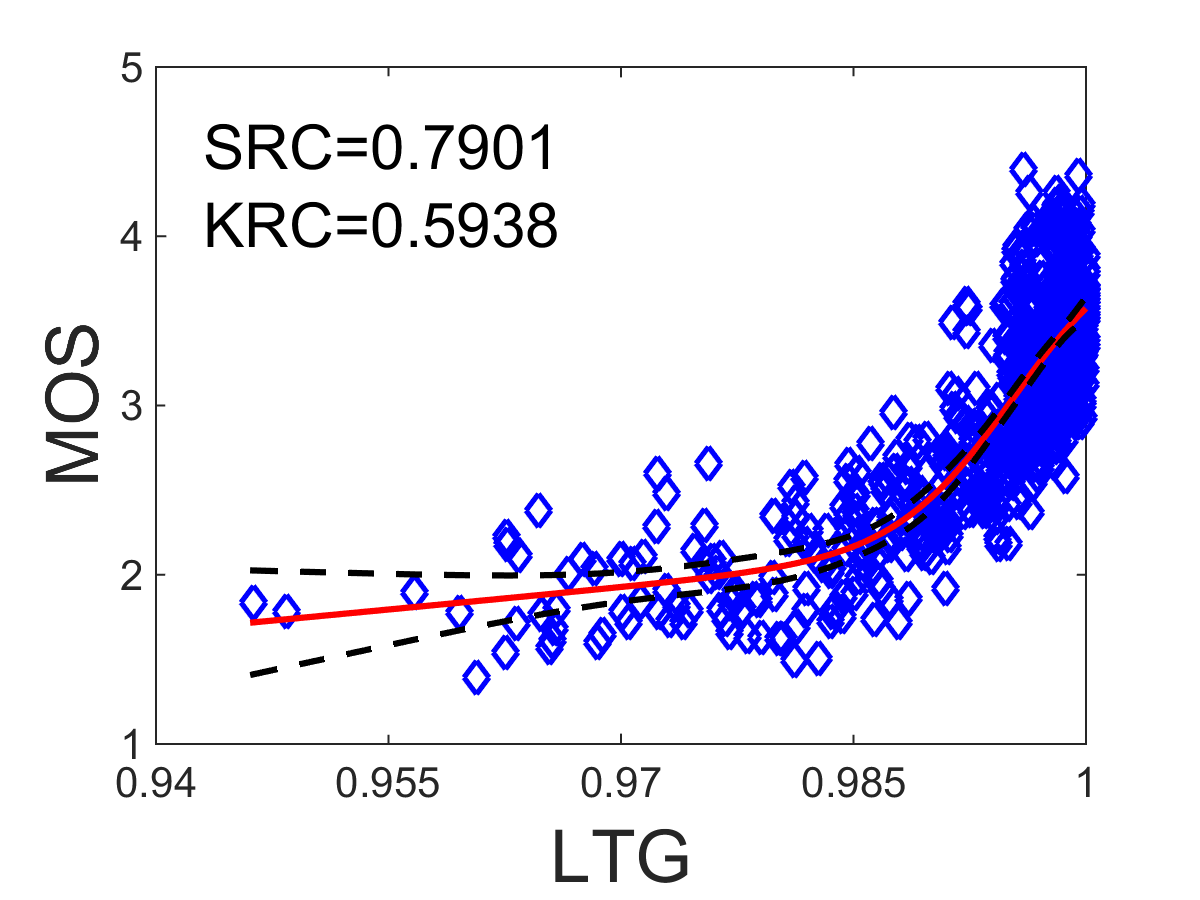}\hspace{-0.2cm}
\includegraphics[height=0.19\linewidth]{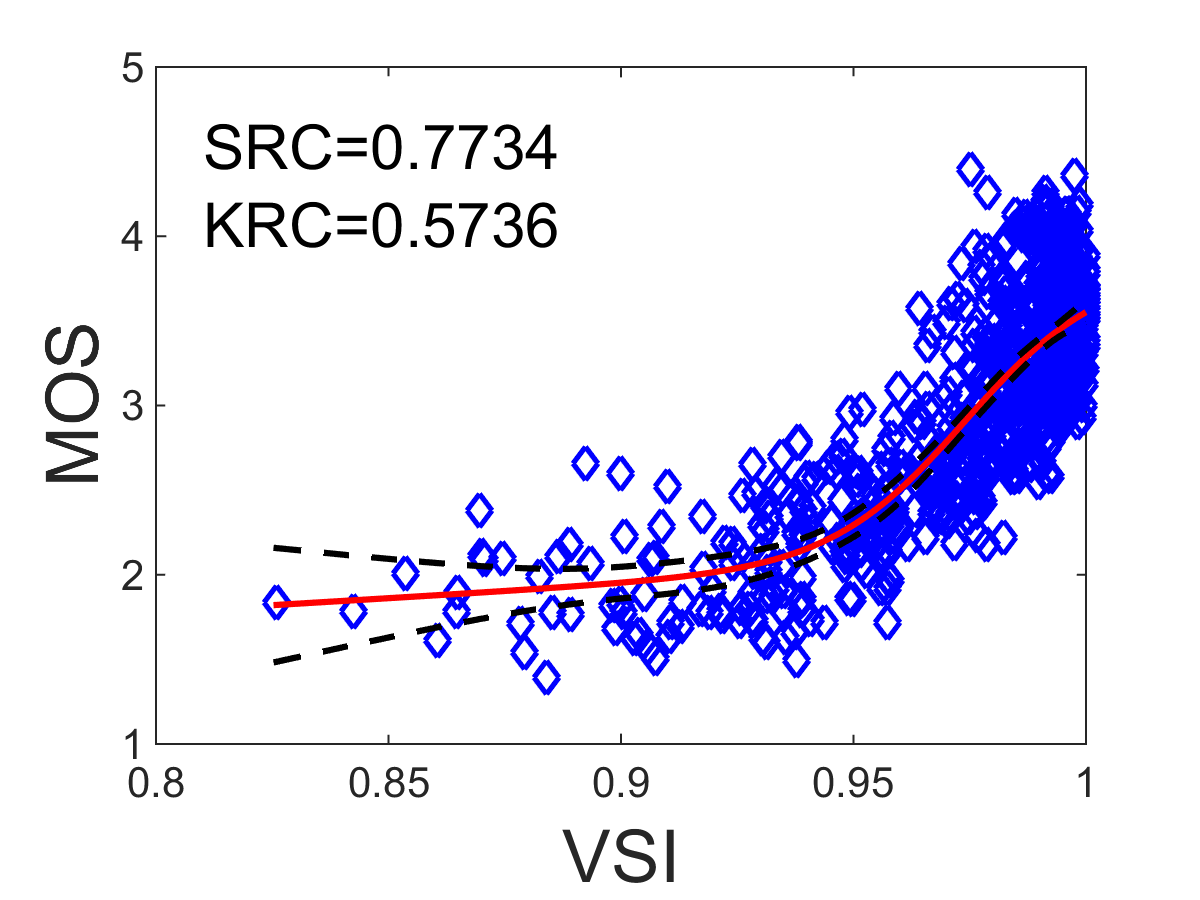}\hspace{-0.2cm}
\includegraphics[height=0.19\linewidth]{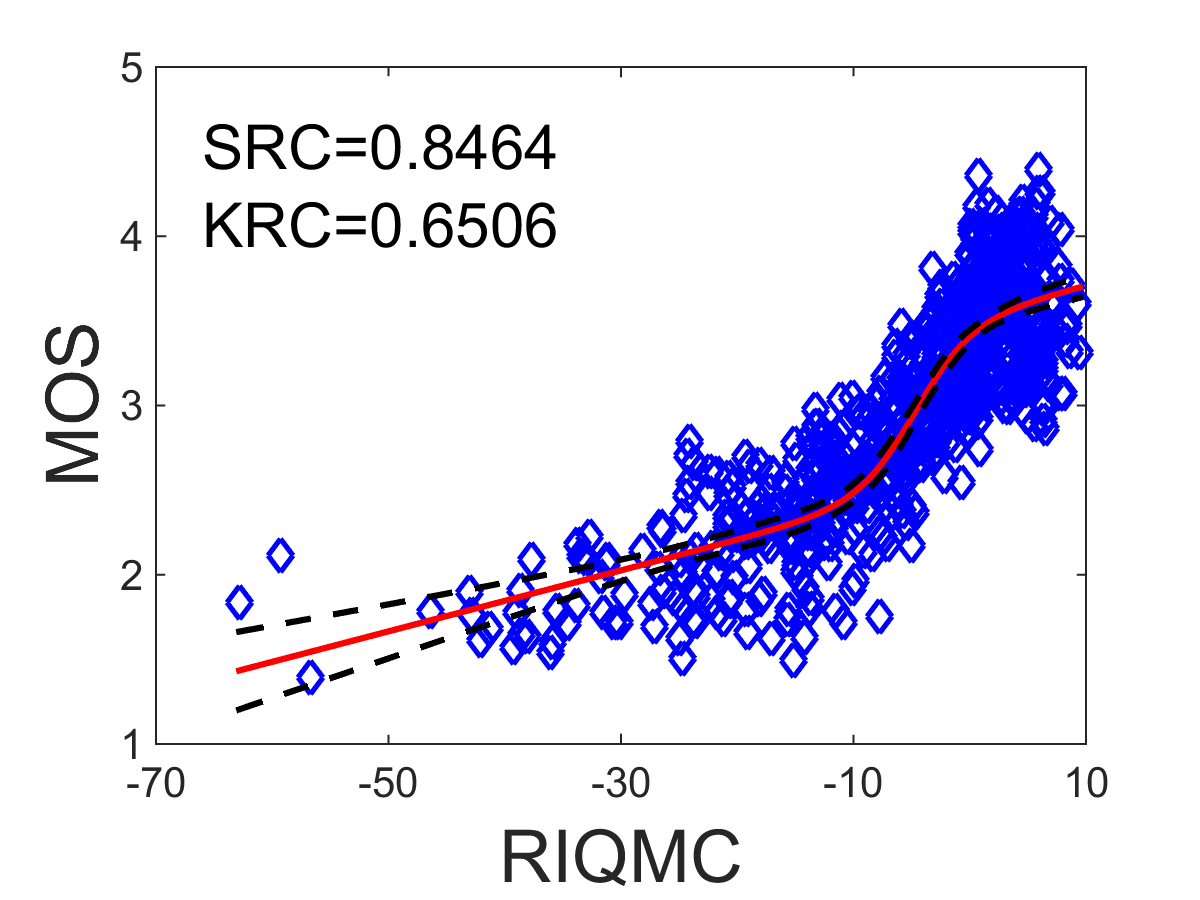}\hspace{-0.2cm}
\includegraphics[height=0.19\linewidth]{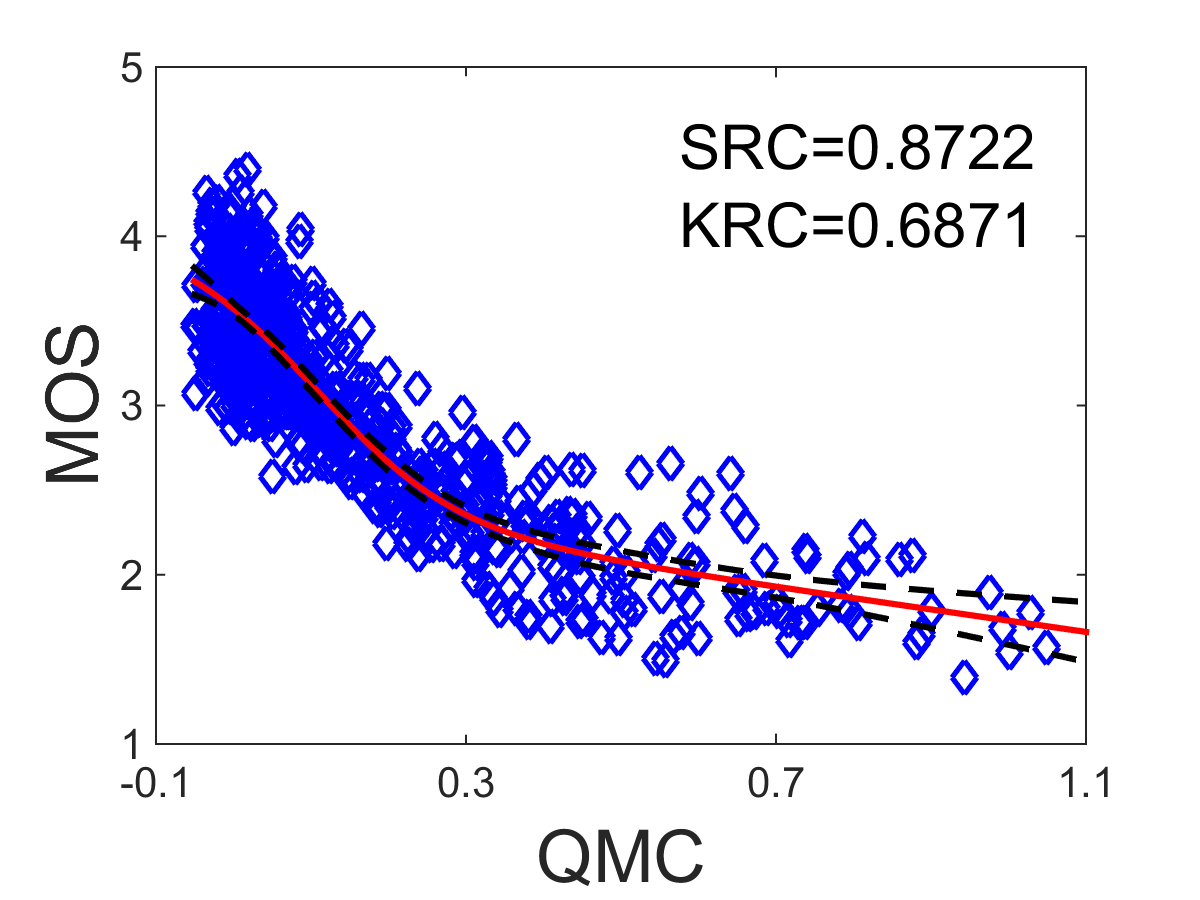}\\
\vspace{0.125cm}
\includegraphics[height=0.19\linewidth]{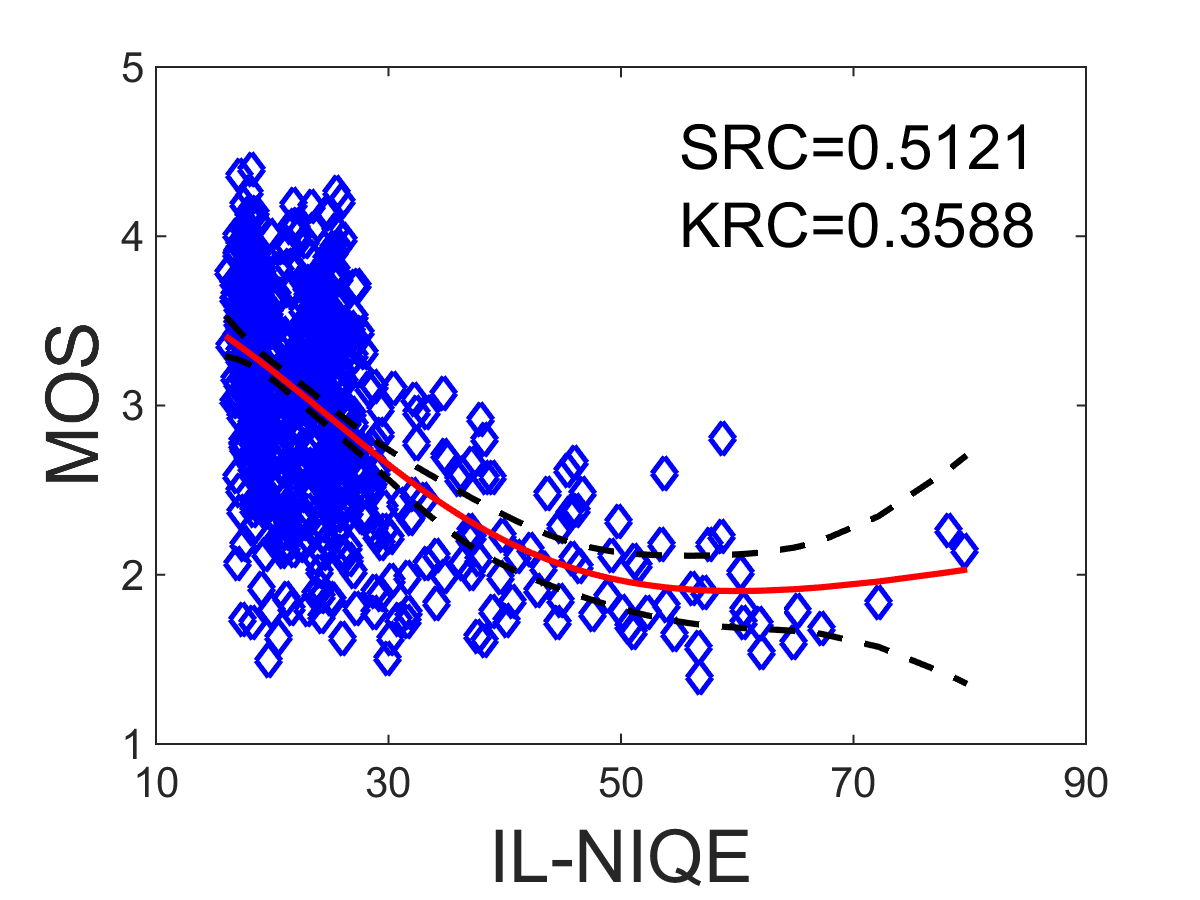}\hspace{-0.2cm}
\includegraphics[height=0.19\linewidth]{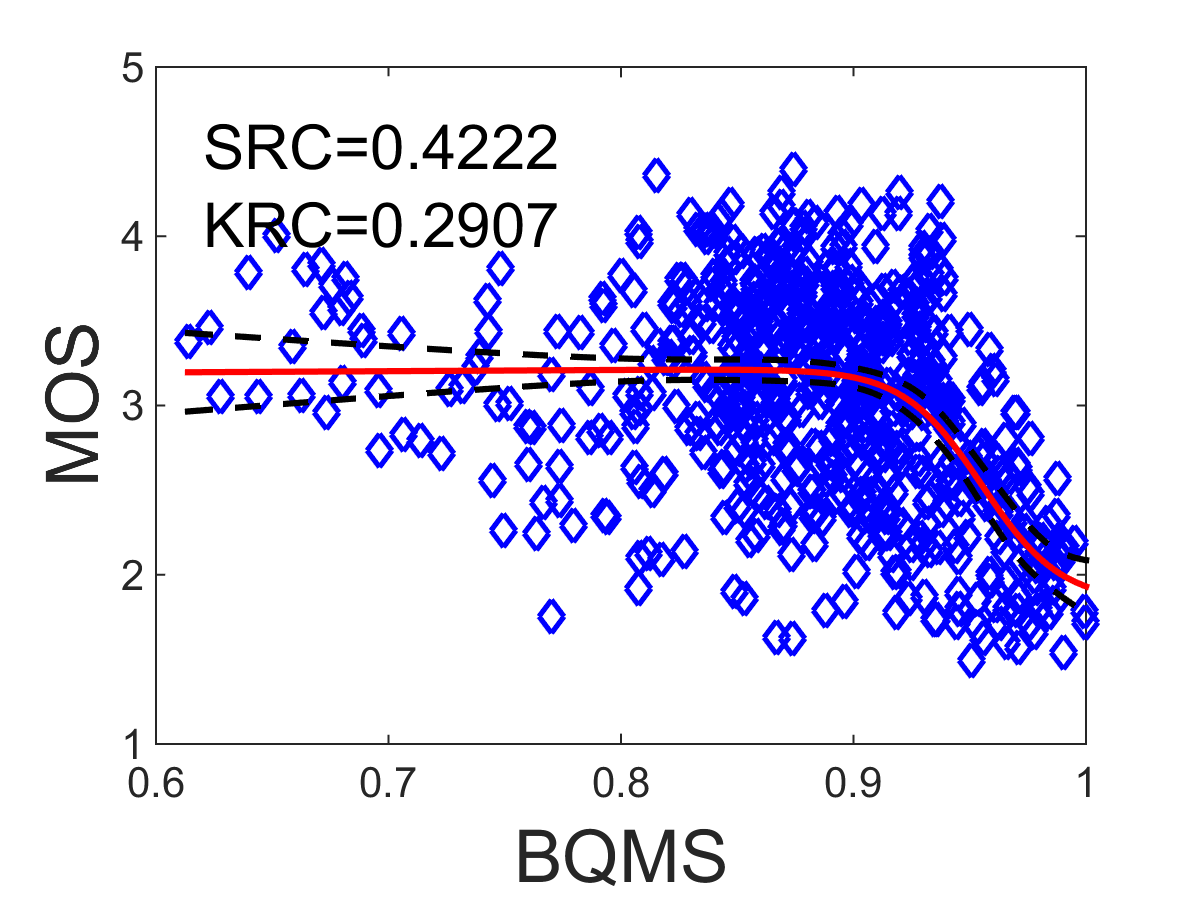}\hspace{-0.2cm}
\includegraphics[height=0.19\linewidth]{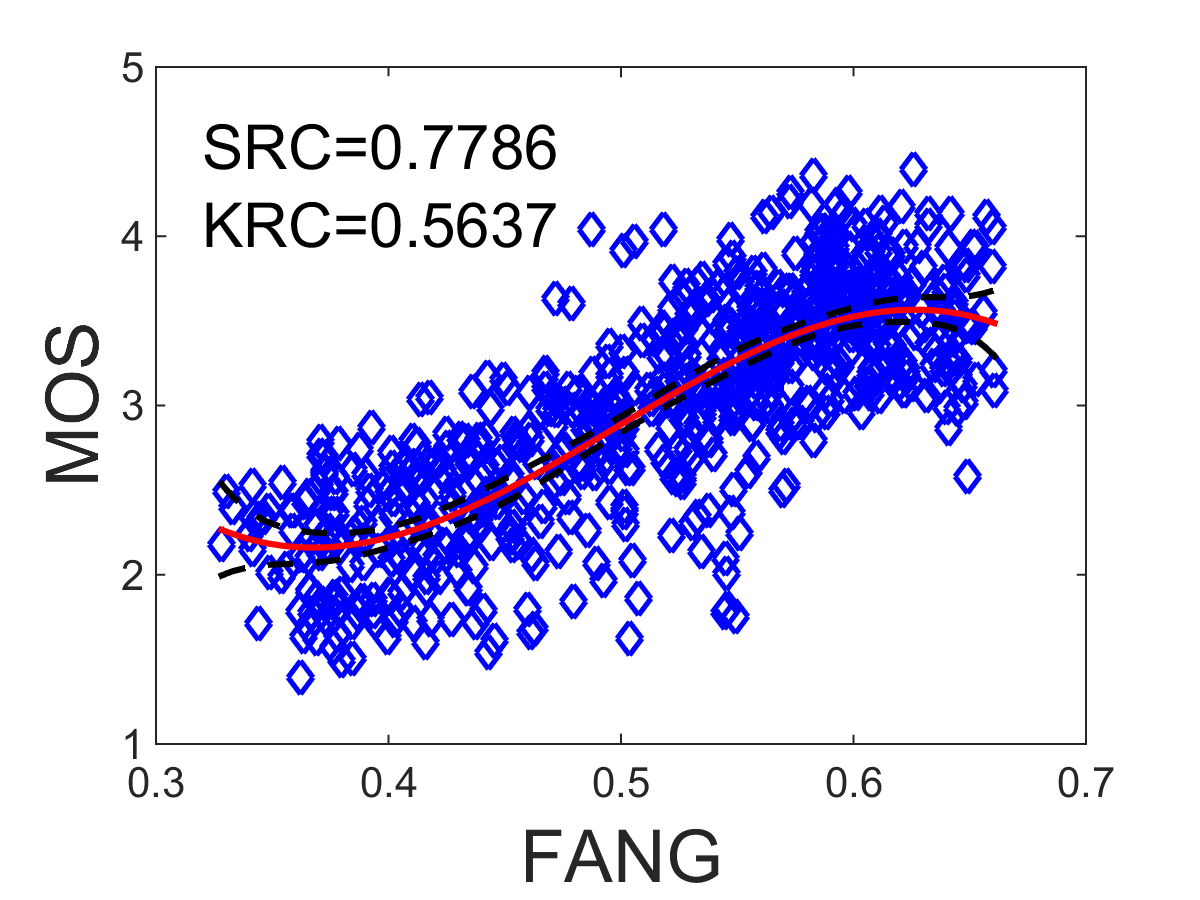}\hspace{-0.2cm}
\includegraphics[height=0.19\linewidth]{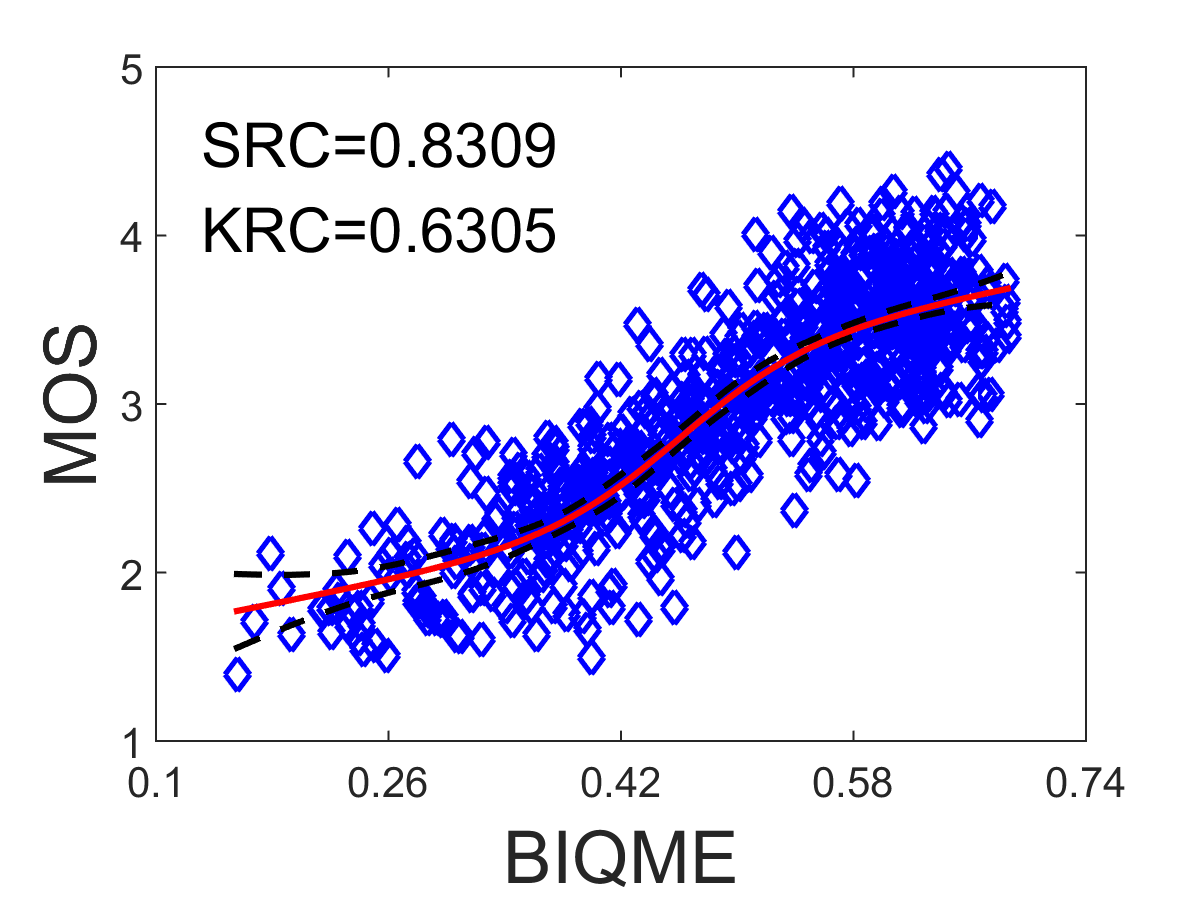}\\
\vspace{0.075cm}
\caption{\small Scatter plots of MOS vs. FR LTG, VSI, RR RIQMC, QMC, and blind IL-NIQE, BQMS, FANG, BIQME on the CCID2014 database. The red lines are curves fitted with the five-parameter logistic function and the black dash lines are 95\% confidence intervals.}
\label{fig:05}
\vspace{0.1cm}
\end{figure*}

Performance Comparison. Most existing NR metrics focus on exploring new effective features, instead of an IQA model. Despite the use of 80\% train-20\% test procedure and leave-one-out cross-validation described in the last subsection, their performance measures are not fair since using only hundreds of training samples to learn the regression module is likely to introduce overfitting. On the other hand, the training and testing data all come from commonly seen datasets in which limited image scenes are included. Clearly, this substantially confines the practical application to a broad scope of visual scenes. In contrast to the opinion-aware blind metrics above, a few opinion-unaware NR-IQA models have been designed upon the NSS regulation [\ref{ref:15}, \ref{ref:19}]. Their modules are trained using about 100 natural images. This paper induces another strategy by using huge amount of training data to learn the regression module, as given in Section \ref{sec:2.2}, and this renders the proposed BIQME an opinion-unaware IQA metric rather than 17 enhancement-related features.

Subsequently, one performance comparison is implemented with opinion-unaware FR, RR and NR quality measures. In this comparison, apart from our BIQME method, we mainly consider the following 13 state-of-the-art IQA models, which encompass: 1) FR FSIM [\ref{ref:06}], LTG [\ref{ref:08}], VSI [\ref{ref:09}], PSIM [\ref{ref:09b}], C-PCQI; 2) RR RRED [\ref{ref:12}], FTQM [\ref{ref:13}], RIQMC [\ref{ref:27}], QMC [\ref{ref:31}]; 3) NR NIQE [\ref{ref:15}], IL-NIQE [\ref{ref:19}], BQMS [\ref{ref:51}], FANG [\ref{ref:33}]\footnote{We deploy the same method and training data in BIQME to learn the regression module of FANG for a fair comparison.}. We have given the results on six datasets in Table \ref{tab:03} and highlighted the best performed metric in each type. Four conclusions can be derived. First, our BIQME metric is obviously superior to other NR-IQA models tested, regardless of general-purpose NQIE and IL-NIQE or distortion-specific BQMS and FANG. Second, the BIQME has acquired an approximating performance to FR C-PCQI and RR RIQMC, which are devised specifically for IQA of contrast alteration under the condition of partial or the whole reference image available, particularly on large-size CID2013 and CCID2014 databases. Third, we surprisingly find that the BIQME metric works effectively on the SIQAD subset; in other words, our BIQME is also fit for assessing the quality of enhanced screen content images. Fourth, compared to opinion-unaware NIQE and IL-NIQE methods which suppose that natural images are of the optimal quality, the proposed opinion-unaware BIQME metric has brought a much better performance, and this gives rise to another strategy in the exploration of opinion-unaware IQA algorithms.

Two mean performance results are included in Table \ref{tab:03} as well. Assuming that the mean index is defined as $\bar{\xi}=\frac{\sum_i\xi_i\cdot \pi_i}{\sum_i\pi_i}$ where $i=\{1,2,...,6\}$ indicates each testing dataset, $\xi_i$ is the performance index on each dataset and $\pi_i$ is the weight, one is the direct mean performance that is computed by setting all the weights to be one, while the other is the weighted mean performance that is computed by assigning the weight $\pi_i$ as the number of images in the testing dataset. One can observe that our blind BIQME technique outclasses all the general-purposed FR-, RR- and NR-IQA methods on average.

In addition to the numerical results, scatter plots of scores between objective IQA approach and subjective opinion are exhibited for straightforward comparison in Fig. \ref{fig:05}, in which the red lines stand for the curves that are fitted by the five-parameter logistic function and the black dash lines stand for 95\% confidence intervals. Besides our NR algorithm, we also include seven competing quality metrics containing FR LTG, VSI, RR RIQMC, QMC, and NR IL-NIQE, BQMS, FANG on the large-scale CCID2014 database for comparison. It is evident that, as compared with those seven IQA approaches considered, our NR BIQME model has given the impressive convergency and monotonicity, noticeably better than blind IL-NIQE, BQMS and FANG metrics.

Runtime Measure. A good IQA model is wished to have high complexity efficiency and low implementation time. So we further compute the runtime of 14 testing IQA methods using the whole 655 images in the CCID2014 database. This experiment is carried out using MATLAB2015 on a desktop computer having 3.20GHz CPU processor and 16GB internal memory. We in table \ref{tab:04} lists the mean runtime of each IQA metric. The proposed BIQME measure, despite using a series computing, only consume less than one second to assess an $768\times 576$ image. Actually, it can be found that each type of features are extracted independently of each other and some features in the same type can be separately calculated (e.g. brightness-related features) when our algorithm runs, so we might introduce the parallel computing strategy to decrease the runtime to a high degree.

\section{Quality-Based Image Enhancement}
\label{sec:4}
Among numerous IQA methods, the majority of them stay at predicting the quality score of an image, yet do not serve to optimize and instruct post-processing techniques towards visual quality improvement. Our BIQME metric, because of its high performance and efficiency, is fit for guiding image enhancement technologies. And moreover, the BIQME works without original references and this makes it apply to many kinds of images, as opposed to some recent works that are only available for enhancing natural images [\ref{ref:55a}, \ref{ref:31}]. Thus we develop a robust BIQME-optimized image enhancement method (BOIEM).

In the BOIEM algorithm, we primarily take into account image brightness and contrast and particularly alter them to a proper level. Enlightened by the RICE enhancement method in [\ref{ref:31}], a two-step framework is constructed. In the first step, we improve two recent enhancement methods, AGCWD [\ref{ref:28}] and RICE [\ref{ref:31}], to successively rectify image brightness and contrast. The AGCWD focuses on weighting the probability density function (PDF) of images by
\begin{equation}
\textrm{PDF}'(z)=\textrm{PDF}_{\max}\bigg(\frac{\textrm{PDF}(z)-\textrm{PDF}_{\min}}{\textrm{PDF}_{\max}-\textrm{PDF}_{\min}}\bigg)^{\lambda_b}
\label{func:15}
\end{equation}
where $z=\{z_{\min},z_{\min}+1,...,z_{\max}\}$; $\textrm{PDF}_{\min}$ and $\textrm{PDF}_{\max}$ respectively indicate the minimum and maximum values in PDF; $\lambda_b$ is a weight parameter. Next, using the weighted PDF to compute the cumulative
distribution function (CDF)
\begin{equation}
\textrm{CDF}'(z)=\sum_{h=0}^z\frac{\textrm{PDF}'(h)}{\sum\textrm{PDF}'}
\label{func:16}
\end{equation}
and produce the enhanced image
\begin{equation}
T(z)=255\bigg(\frac{z}{255}\bigg)^{1-\textrm{CDF}'(z)}.
\label{func:17}
\end{equation}
In [\ref{ref:28}], the weight parameter $\lambda_b$ is empirically assigned as a constant number. But it was found that this parameter value sometimes leads to over-enhancement, making the processed images excessively brilliant [\ref{ref:27}].

The RICE offers a more complete histogram modification framework to be optimized by quality metric. In RICE, it is hypothesized that the ideal histogram of properly enhanced images is towards having uniform PDF, close to the original histogram, and of positively skewed statistics to elevate the surface quality [\ref{ref:56}]. Based on this hypothesis, an optimization function was established:
\begin{equation}
\tilde{\mathbf{h}}=\mathop{\textrm{minimize}}_{\mathbf{h}}\|\mathbf{h}-\mathbf{h}_\mathbf{i}\|+\lambda_e\|\mathbf{h}-\mathbf{h}_\mathbf{e}\|+\lambda_s\|\mathbf{h}-\mathbf{h}_\mathbf{s}\|
\label{func:18}
\end{equation}
where $\mathbf{h}_\mathbf{i}$, $\mathbf{h}_\mathbf{e}$ and $\mathbf{h}_\mathbf{s}$ are histograms of uniform distribution, original distribution and positively skewed statistics; $\lambda_e$ and $\lambda_s$ are weighting parameters to be ascertained. Through some simplifications, an analytical solution was derived:
\begin{equation}
\tilde{\mathbf{h}}=\frac{\mathbf{h}_\mathbf{i}+\lambda_e\mathbf{h}_\mathbf{e}+\lambda_s\mathbf{h}_\mathbf{s}}{1+\lambda_e+\lambda_s}.
\label{func:19}
\end{equation}
Given the output histogram $\tilde{\mathbf{h}}$, the histogram matching and quality-optimized techniques are used for enhancing images. Notice that two weights $\lambda_e$ and $\lambda_s$ are adaptively determined by quality metric on three pairs of parameter candidates, and therefore the RICE algorithm is good at enhancing natural images. Nonetheless, it fails for other types of images, such as low-light images, because the RICE method do not adjust brightness and moreover it requires reference images in the quality-based optimization.

\begin{figure}[!t]
\center
\small
\vspace{0.05cm}
\includegraphics[width=2.1cm]{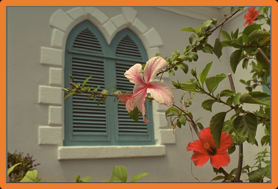} \hspace{-0.1cm}
\includegraphics[width=2.1cm]{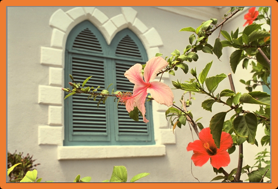} \hspace{-0.1cm}
\includegraphics[width=2.1cm]{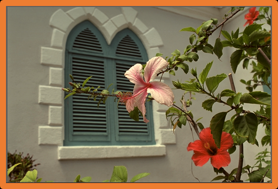}  \hspace{-0.1cm}
\includegraphics[width=2.1cm]{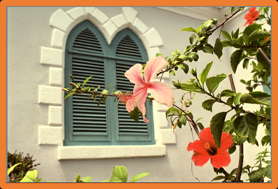} \\
\vspace{0.125cm}
\includegraphics[width=2.1cm]{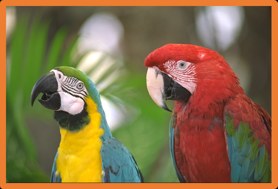} \hspace{-0.1cm}
\includegraphics[width=2.1cm]{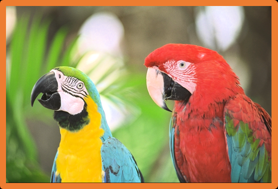} \hspace{-0.1cm}
\includegraphics[width=2.1cm]{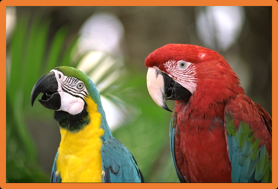}  \hspace{-0.1cm}
\includegraphics[width=2.1cm]{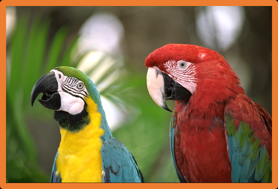} \\
\vspace{0.125cm}
\includegraphics[width=2.1cm]{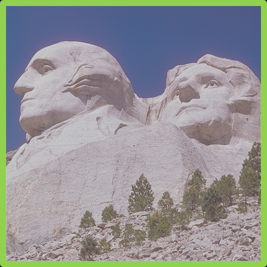} \hspace{-0.1cm}
\includegraphics[width=2.1cm]{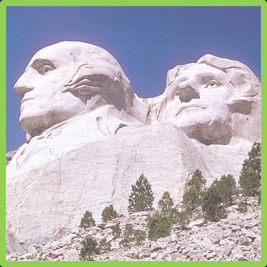} \hspace{-0.1cm}
\includegraphics[width=2.1cm]{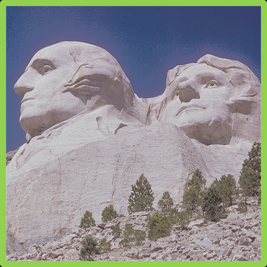}  \hspace{-0.1cm}
\includegraphics[width=2.1cm]{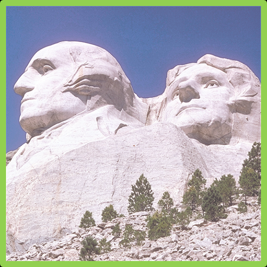} \\
\vspace{0.125cm}
\includegraphics[width=2.1cm]{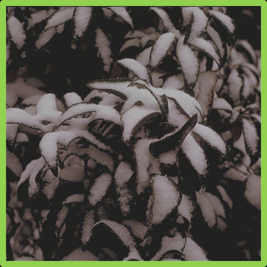} \hspace{-0.1cm}
\includegraphics[width=2.1cm]{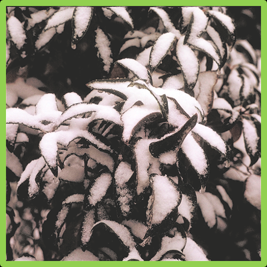} \hspace{-0.1cm}
\includegraphics[width=2.1cm]{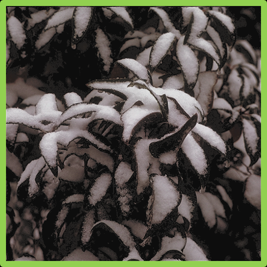}  \hspace{-0.1cm}
\includegraphics[width=2.1cm]{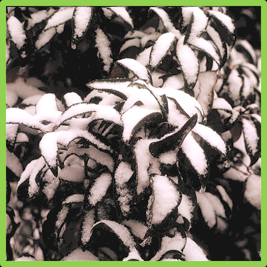} \\
\vspace{0.125cm}
\includegraphics[width=2.1cm]{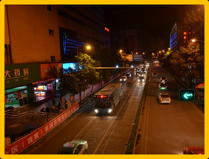} \hspace{-0.1cm}
\includegraphics[width=2.1cm]{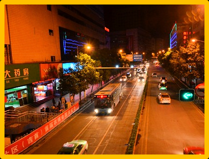} \hspace{-0.1cm}
\includegraphics[width=2.1cm]{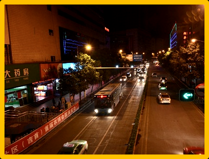}  \hspace{-0.1cm}
\includegraphics[width=2.1cm]{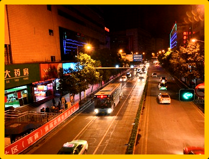} \\
\vspace{0.125cm}
\includegraphics[width=2.1cm]{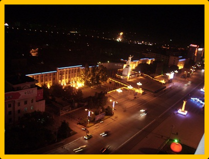} \hspace{-0.1cm}
\includegraphics[width=2.1cm]{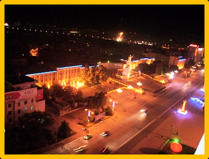} \hspace{-0.1cm}
\includegraphics[width=2.1cm]{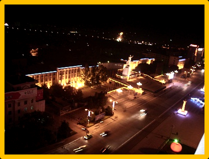}  \hspace{-0.1cm}
\includegraphics[width=2.1cm]{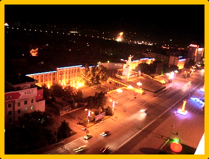} \\
\vspace{0.125cm}
\includegraphics[width=2.1cm]{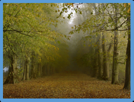} \hspace{-0.1cm}
\includegraphics[width=2.1cm]{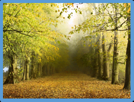} \hspace{-0.1cm}
\includegraphics[width=2.1cm]{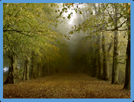}  \hspace{-0.1cm}
\includegraphics[width=2.1cm]{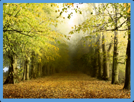} \\
\vspace{0.125cm}
\includegraphics[width=2.1cm]{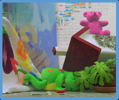} \hspace{-0.1cm}
\includegraphics[width=2.1cm]{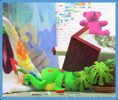} \hspace{-0.1cm}
\includegraphics[width=2.1cm]{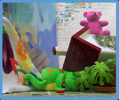}  \hspace{-0.1cm}
\includegraphics[width=2.1cm]{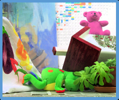} \\
\flushleft
\quad\;\;\,\,Original\quad\;\;\;\;\,\,AGCWD [\ref{ref:28}]\qquad\,\!RICE [\ref{ref:31}]\quad\;\;\;BOIEM (Pro.)
\caption{\small Comparison of image enhancement technologies on natural images, low-contrast images, low-light images and dehazed images.}
\label{fig:06}
\end{figure}

In the design of our BOIEM model, a cascade of modified AGCWD and RICE are utilized with parameters ($\lambda_b$, $\lambda_s$ and $\lambda_e$) to be decided in the first step. Then the proposed blind BIQME algorithm is used to optimize these three parameters:
\begin{equation}
\lambda_b,\lambda_s,\lambda_e=\mathop{\textrm{maximize}}_{\lambda_b,\lambda_s,\lambda_e}Q_B(T_R[T_A(\mathbf{s},\lambda_b),\lambda_s,\lambda_e])
\label{func:20}
\end{equation}
where $Q_B$, $T_R$ and $T_A$ are respectively associated to BIQME, RICE and AGCWD. Thereafter, we exploit these parameters to enhance images. By extensive experiments, it was observed that the images enhanced by simultaneously optimizing three parameters and separately optimizing the former $\lambda_b$ and the latter two $\lambda_s$ and $\lambda_e$ look almost the same. So, following the speed-up strategy applied in [\ref{ref:31}], the BOIEM only conducts six times BIQME for optimization, the first three times to enumerate three candidates $\{0.3,0.5,0.7\}$ to pick the best $\lambda_b$ for image brightness rectification and the latter three times to pick the optimal $\lambda_s$ and $\lambda_e$ from candidates given in [\ref{ref:31}] for image contrast improvement. In accordance to the selected parameters, we can finally generate the enhanced images.

Through careful rectification of brightness and contrast and quality-guided optimization, the proposed BOIEM model can well enhance natural images, low-contrast images, low-light images and dehazed images. Part of results are illustrated in Fig. \ref{fig:06}. Images circled with red, green, orange and blue rectangles are separately natural images, low-contrast images, low-light images and dehazed images [\ref{ref:29}]. More results can be found in the supplementary file C. Two lately developed enhancement techniques, AGCWD [\ref{ref:28}] and RICE [\ref{ref:31}], are included for comparison, as shown in Fig. \ref{fig:06}.

In contrast, using the fixed weighting number $\lambda_b$, AGCWD often introduces over-brightness, especially for natural images which themselves have appropriate luminance. Furthermore, there lacks the procedure of contrast gain in AGCWD and this makes details hard to appear. Seeing the third column, RICE shows its good ability to enhance natural images, like erasing a curtain of fog from photos. Yet RICE is helpless for low-light images, which is very possibly because there is no luminance alteration term in (\ref{func:19}), and on the other hand it regards the input image as a high-quality natural image in the IQA-based optimization towards further improving the visual quality of input images. By systematically incorporating these two good enhancement technologies and high-performance blind BIQME algorithm to optimize parameters, one can see in the rightmost column in Fig. \ref{fig:06} that the proposed BOIEM algorithm is able to well enhance natural images, low-contrast images, low-light images and dehazed images, which makes them have suitable brightness and contrast and display more details.

\section{Conclusion}
\label{sec:5}
In this paper we have constructed a general framework for quality assessment of enhanced images and its application to robust enhancement technologies. As for an enhanced image, we take into consideration five influencing factors: image contrast, sharpness, brightness, colorfulness and naturalness, and associated 17 features to blindly predict its visual quality. Thorough experiments using three categories of performance comparison strategies demonstrate that the proposed BIQME metric is remarkably superior to the same type of NR-IQA methods using nine relevant image datasets. In comparison to FR and RR algorithms, our BIQME metric implements better than general-purpose FR- and RR-IQA methods, but slightly inferior to those FR and RR quality measures dedicated to IQA of contrast change. It deserves the stress that on one hand each type of features used in BIQME is independent of others, so we might usher parallel computing to increase its computational efficiency to some extent, and on the other hand our IQA framework is flexible in inducing novel features to derive higher performance.

With the blind BIQME metric for optimization, we have devised a framework rectifying image brightness and contrast successively, to properly enhance natural images, low-contrast images, low-light images and dehazed images. It is worthy to mention that incorporating more procedures, such as image haze removal, will make our enhancement framework more universal.

Visual saliency is an intrinsic attribute of the human visual system and this renders a possible future work by conducting saliency detection methods to modify brightness-, sharpness- and colorfulness-related features towards better performance. As compared to existing opinion-unaware NR-IQA methods, our IQA framework provides a new strategy in the design of opinion-unaware blind quality measures, particularly for complicated distortions such as image dehazing. So another feature work might turn to convert/extend our framework to blind IQA tasks of denoising, deblurring and super-resolution with new relevant features injected.


\begin{thebibliography}{1}
\bibitem{01}\label{ref:01}
H. R. Sheikh, Z. Wang, L. Cormack, and A. C. Bovik, ``LIVE image quality assessment Database Release 2,''  2006, Online at: http://live.ece.utexas.edu/research/quality

\bibitem{02}\label{ref:02}
K. Gu, G. Zhai, X. Yang, and W. Zhang, ``Hybrid no-reference quality metric for singly and multiply distorted images,'' \emph{IEEE Trans. Broadcasting}, vol. 60, no. 3, pp. 555-567, Sept. 2014.

\bibitem{03}\label{ref:03}
K. Gu, M. Liu, G. Zhai, X. Yang, and W. Zhang, ``Quality assessment considering viewing distance and image resolution,'' \emph{IEEE Trans. Broadcasting}, vol. 61, no. 3, pp. 520-531, Sept. 2015.

\bibitem{41}\label{ref:41}
C. Li, A. C. Bovik, and X. Wu, ``Blind image quality assessment using a general regression neural network,'' \emph{IEEE Trans. Neural Networks}, vol. 22, no. 5, pp. 793-399, May 2011.

\bibitem{05a}\label{ref:05a}
M. Narwaria and W. Lin, ``Objective image quality assessment based on support vector regression,'' \emph{IEEE Trans. Neural Networks}, vol. 21, no. 3, pp. 515-519, Mar. 2010.

\bibitem{21}\label{ref:21}
S. Wang, K. Gu, S. Ma, W. Lin, X. Liu, and W. Gao, ``Guided image contrast enhancement based on retrieved images in cloud,'' \emph{IEEE Trans. Multimedia}, vol. 18, no. 2, pp. 219-232, Feb. 2016.

\bibitem{05}\label{ref:05}
Z. Wang, A. C. Bovik, H. R. Sheikh, and E. P. Simoncelli, ``Image quality assessment: From error visibility to structural similarity,'' \emph{IEEE Trans. Image Process.}, vol. 13, no. 4, pp. 600-612, Apr. 2004.

\bibitem{06}\label{ref:06}
L. Zhang, L. Zhang, X. Mou, and D. Zhang, ``FSIM: A feature similarity index for image quality assessment,'' \emph{IEEE Trans. Image Process.}, vol. 20, no. 8, pp. 2378-2386, Aug. 2011.

\bibitem{08}\label{ref:08}
K. Gu, G. Zhai, X. Yang, and W. Zhang, ``An efficient color image quality metric with local-tuned-global model,'' \emph{in Proc. IEEE Int. Conf. Image Process.}, pp. 506-510, Oct. 2014.

\bibitem{09}\label{ref:09}
L. Zhang, Y. Shen, and H. Li, ``VSI: A visual saliency induced index for perceptual image quality assessment,'' \emph{IEEE Trans. Image Process.}, vol. 23, no. 10, pp. 4270-4281, Oct. 2014.

\bibitem{09b}\label{ref:09b}
K. Gu, L. Li, H. Lu, X. Min, and W. Lin, ``A fast reliable image quality predictor by fusing micro- and macro-structures,'' \emph{IEEE Trans. Ind. Electron.}, 2017, to appear.

\bibitem{09a}\label{ref:09a}
X. Gao, W. Lu, X. Li, and D. Tao, ``Wavelet-based contourlet in quality evaluation of digital images,'' \emph{Neurocomputing}, vol. 72, no. 1, pp. 378-385, Dec. 2008.

\bibitem{10}\label{ref:10}
D. Tao, X. Li, W. Lu, and X. Gao, ``Reduced-reference IQA in contourlet domain,'' \emph{IEEE Trans. Syst., Man, Cybern. B, Cybern.}, vol. 39, no. 6, pp. 1623-1726, Dec. 2009.

\bibitem{11}\label{ref:11}
X. Gao, W. Lu, D. Tao, and X. Li, ``Image quality assessment based on multiscale geometric analysis,'' \emph{IEEE Trans. Image Process.}, vol. 18, no. 7, pp. 1409-1423, Jul. 2009.

\bibitem{12}\label{ref:12}
R. Soundararajan and A. C. Bovik, ``RRED Indices: Reduced Reference Entropic Differencing for Image Quality Assessment,'' \emph{IEEE Trans. Image Process.}, vol. 21, no. 2, pp. 517-526, Feb. 2012.

\bibitem{13}\label{ref:13}
M. Narwaria, W. Lin, I. V. McLoughlin, S. Emmanuel, and L. T. Chia, ``Fourier transform-based scalable image quality measure,'' \emph{IEEE Trans. Image Process.}, vol. 21, no. 8, pp. 3364-3377, Aug. 2012.

\bibitem{14}\label{ref:14}
A. Mittal, A. K. Moorthy, and A. C. Bovik, ``No-reference image quality assessment in the spatial domain,'' \emph{IEEE Trans. Image Process.}, pp. 4695-4708, vol. 21, no. 12, Dec. 2012.

\bibitem{17}\label{ref:17}
K. Gu, G. Zhai, X. Yang, and W. Zhang, ``Using free energy principle for blind image quality assessment,'' \emph{IEEE Trans. Multimedia}, vol. 17, no. 1, pp. 50-63, Jan. 2015.

\bibitem{15}\label{ref:15}
A. Mittal, R. Soundararajan, and A. C. Bovik, ``Making a `completely blind' image quality analyzer,'' \emph{IEEE Sig. Process. Lett.}, pp. 209-212, vol. 22, no. 3, Mar. 2013.

\bibitem{19}\label{ref:19}
L. Zhang, L. Zhang, and A. C. Bovik, ``A feature-enriched completely blind image quality evaluator,'' \emph{IEEE Trans. Image Process.}, vol. 24, no. 8, pp. 2579-2591, Aug. 2015.

\bibitem{15a}\label{ref:15a}
R. A. Manap, L. Shao, and A. F. Frangi, ``Non-parametric quality assessment of natural images,'' \emph{IEEE Multimedia}, 2016, in press.

\bibitem{24}\label{ref:24}
N. Ponomarenko \emph{et al.}, ``TID2008-A database for evaluation of full-reference visual quality assessment metrics,'' \emph{Advances of Modern Radioelectronics}, vol. 10, pp. 30-45, 2009.

\bibitem{25}\label{ref:25}
E. C. Larson and D. M. Chandler, ``Most apparent distortion: Full-reference image quality assessment and the role of strategy,'' \emph{J. Electr. Imag.}, vol. 19, no. 1, Mar. 2010. Online at: http://vision.okstate.edu/csiq

\bibitem{26}\label{ref:26}
N. Ponomarenko \emph{et al.}, ``Image database TID2013: Peculiarities, results and perspectives,'' \emph{Sig. Process.: Image Commun.}, vol. 30, pp. 57-55, Jan. 2015.

\bibitem{16}\label{ref:16}
X. Gao, F. Gao, D. Tao, and X. Li, ``Universal blind image quality assessment metrics via natural scene statistics and multiple kernel learning,'' \emph{IEEE Trans. Neural Netw. Learning Syst.}, vol. 24, no. 12, pp. 2013-2026, Dec. 2013.

\bibitem{18}\label{ref:18}
W. Hou, X. Gao, D. Tao, and X. Li, ``Blind image quality assessment via deep learning,'' \emph{IEEE Trans. Neural Netw. Learning Syst.}, vol. 26, no. 6, pp. 1275-1286, Jun. 2015.

\bibitem{18a}\label{ref:18a}
F. Shao, W. Tian, W. Lin, G. Jiang, and Q. Dai, ``Toward a blind deep quality evaluator for stereoscopic images based on monocular and binocular interactions,'' \emph{IEEE Trans. on Image Process.}, vol. 25, no. 5, pp. 2059-2074, Mar. 2016.

\bibitem{20}\label{ref:20}
F. Gao, D. Tao, X. Gao, and X. Li, ``Learning to rank for blind image quality assessment,'' \emph{IEEE Trans. Neural Netw. Learning Syst.}, vol. 26, no. 10, pp. 2275-2290, Oct. 2015.

\bibitem{37}\label{ref:37}
D. L. Ruderman, ``The statistics of natural images,'' \emph{Netw. Comput. Neural Syst.}, vol. 5, no. 4, pp. 517-548, 1994.

\bibitem{27}\label{ref:27}
K. Gu, G. Zhai, W. Lin, and M. Liu, ``The analysis of image contrast: From quality assessment to automatic enhancement,'' \emph{IEEE Trans. Cybernetics}, vol. 46, no. 1, pp. 284-297, Jan. 2016.

\bibitem{28}\label{ref:28}
S.-C. Huang, F.-C. Cheng, and Y.-S. Chiu, ``Efficient contrast enhancement using adaptive gamma correction with weighting distribution,'' \emph{IEEE Trans. Image Process.}, pp. 1032-1041, vol. 22, no. 3, Mar. 2013.

\bibitem{29}\label{ref:29}
K. He, J. Sun, and X. Tang, ``Single image haze removal using dark channel prior,'' \emph{IEEE Trans. Pattern Anal. Mach. Intell.}, vol. 33, no. 12, pp. 2341-2353, Dec. 2011.

\bibitem{30}\label{ref:30}
K. Gu, G. Zhai, X. Yang, W. Zhang, and M. Liu, ``Subjective and objective quality assessment for images with contrast change,'' \emph{in Proc. IEEE Int. Conf. Image Process.}, pp. 383-387, Sep. 2013.

\bibitem{31}\label{ref:31}
K. Gu, G. Zhai, X. Yang, W. Zhang, and C. W. Chen, ``Automatic contrast enhancement technology with saliency preservation,'' \emph{IEEE Trans. Circuits Syst. Video Technol.}, vol. 25, no. 9, pp. 1480-1494, Sept. 2015.

\bibitem{32}\label{ref:32}
S. Wang, K. Ma, H. Yeganeh, Z. Wang and W. Lin, ``A patch-structure representation method for quality assessment of contrast changed images,'' \emph{IEEE Sig. Process. Lett.}, pp. 2387-2390, vol. 22, no. 7, Dec. 2015.

\bibitem{33}\label{ref:33}
Y. Fang, K. Ma, Z. Wang, W. Lin, Z. Fang, and G. Zhai, ``No-reference quality assessment of contrast-distorted images based on natural scene statistics,'' \emph{IEEE Sig. Process. Lett.}, vol. 22, no. 7, pp. 838-842, Jul. 2015.

\bibitem{34}\label{ref:34}
Z. Chen, T. Jiang, and Y. Tian, ``Quality assessment for comparing image enhancement algorithms,'' \emph{in Proc. IEEE Conf. Comput. Vis. and Pattern Recognit.}, pp. 3003-3010, Jun. 2014.

\bibitem{35}\label{ref:35}
A. Oliva and A. Torralba, ``Modeling the shape of the scene: A holistic representation of the spatial envelope,'' \emph{Int. J. Comput. Vis.}, vol. 42, no. 3, pp. 145-175, May 2001.

\bibitem{36}\label{ref:36}
M. A. Stricker and M. Orengo, ``Similarity of color images,'' \emph{in IS\&T/SPIE's Symposium on Electronic Imaging: Science \& Technology. International Society for Optics and Photonics}, pp. 381-392, Mar. 1995.

\bibitem{37a}\label{ref:37a}
https://en.wikipedia.org/wiki/Contrast\_(vision)

\bibitem{38}\label{ref:38}
A. V. Oppenheim and J. S. Lim, ``The importance of phase in signals,'' \emph{Proc. IEEE}, vol. 69, no. 5, pp. 529-541, Nov. 1981.

\bibitem{39}\label{ref:39}
M. C. Morrone, J. Ross, D. C. Burr, and R. Owens, ``Mach bands are phase dependent,'' \emph{Nature}, vol. 324, no. 6049, pp. 250-253, Nov. 1986.

\bibitem{40}\label{ref:40}
P. Kovesi, ``Image features from phase congruency,'' \emph{Videre: J. Comp. Vis. Res.}, vol. 69, no. 3, pp. 1-26, 1999.

\bibitem{42}\label{ref:42}
I. I. A. Groen, S. Ghebreab, H. Prins, V. A. F. Lamme, and H. S. Scholte, ``From image statistics to scene gist: Evoked neural activity reveals transition from low-level natural image structure to scene category,'' \emph{J. Neurosci.}, vol. 33, no. 48, pp. 18814-18824, Nov. 2013.

\bibitem{43}\label{ref:43}
H. S. Scholte, S. Ghebreab, L. Waldorp, A. W. Smeulders, and V. A. Lamme, ``Brain responses strongly correlate with Weibull image statistics when processing natural images,'' \emph{J. Vis.}, vol. 9, no. 4, pp. 1-15, Apr. 2009.

\bibitem{44}\label{ref:44}
D. J. Heeger, ``Normalization of cell responses in cat striate cortex,'' \emph{Vis. Neurosci.}, vol. 9, no. 2, pp. 181-197, 1992.

\bibitem{45}\label{ref:45}
L. K. Choi, J. You, and A. C. Bovik, ``Referenceless prediction of perceptual fog density and perceptual image defogging,'' \emph{IEEE Trans. Image Process.}, vol. 24, no. 11, pp. 3888-3901, Nov. 2015.

\bibitem{45a}\label{ref:45a}
D. Hasler and S. E. Suesstrunk, ``Measuring colorfulness in natural images,'' \emph{Proc. SPIE}, vol. 5007, pp. 87-95, Jun. 2003.

\bibitem{46}\label{ref:46}
E. Kee and H. Farid, ``A perceptual metric for photo retouching,'' \emph{Proceedings of the National Academy of Sciences of the United States of America (PNAS)}, vol. 108, no. 50, pp. 19907-19912, Dec. 2011.

\bibitem{46a}\label{ref:46a}
C. Vu, T. Phan, and D. Chandler, ``S$_\textrm{3}$: A spectral and spatial measure of local perceived sharpness in natural images,'' \emph{IEEE Trans. Image Process.}, vol. 21, no. 3, pp. 934-945, Mar. 2012.

\bibitem{47}\label{ref:47}
P. V. Vu and D. M. Chandler, ``A fast wavelet-based algorithm for global and local image sharpness estimation,'' \emph{IEEE Sig. Process. Lett.}, vol. 19, no. 7, pp. 423-426, Jul. 2012.

\bibitem{48}\label{ref:48}
K. Gu, G. Zhai, W. Lin, X. Yang, and W. Zhang, ``No-reference image sharpness assessment in autoregressive parameter space,'' \emph{IEEE Trans. Image Process.}, vol. 24, no. 10, pp. 3218-3231, Oct. 2015.

\bibitem{49}\label{ref:49}
K. Gu, S. Wang, G. Zhai, S. Ma, X. Yang, W. Lin, W. Zhang, and W. Gao, ``Blind quality assessment of tone-mapped images via analysis of information, naturalness and structure,'' \emph{IEEE Trans. Multimedia}, vol. 18, no. 3, pp. 432-443, Mar. 2016.

\bibitem{50}\label{ref:50}
S. T. Roweis and L. K. Saul, ``Nonlinear dimensionality reduction by locally linear embedding,'' \emph{Science}, vol. 290, no. 5500, pp. 2323-2326, 2000.

\bibitem{51}\label{ref:51}
K. Gu, G. Zhai, W. Lin, X. Yang, and W. Zhang, ``Learning a blind quality evaluation engine of screen content images,'' \emph{Neurocomputing}, vol. 196, pp. 140-149, Jul. 2016.

\bibitem{51a}\label{ref:51a}
R. Datta, D. Joshi, J. Li, and J. Z. Wang, ``Studying aesthetics in photographic images using a computational approach,'' \emph{in Proc. Eur. Conf. Comput. Vis.}, pp. 288-301, May 2006.

\bibitem{52}\label{ref:52}
D. Martin, C. Fowlkes, D. Tal, and J. Malik, ``A database of human segmented natural images and its application to evaluating segmentation algorithms and measuring ecological statistics,'' \emph{in Proc. IEEE Int. Conf. Comput. Vis.}, pp. 416-423, 2001.

\bibitem{53}\label{ref:53}
X. Tang, W. Luo, and X. Wang, ``Content-based photo quality assessment,'' \emph{IEEE Trans. Multimedia}, vol. 15, no. 8, pp. 1930-1943, Dec. 2013.

\bibitem{54}\label{ref:54}
C-C. Chang and C-J. Lin, ``LIBSVM: a library for support vector machines,'' \emph{ACM Trans. Intelligent Systems and Technology}, vol. 2, no. 3, 2011, Online at: http://www.csie.ntu.edu.tw/$\sim$cjlin/libsvm

\bibitem{54a}\label{ref:54a}
Z. Wang, S. Chang, F. Dolcos, D. Beck, D. Liu, and T. S. Huang, ``Brain-inspired deep networks for image aesthetics assessment,'' \emph{arXiv preprint arXiv:1601.04155}, 2016.

\bibitem{54b}\label{ref:54b}
G. Hinton, O. Vinyals, and J. Dean, ``Distilling the knowledge in a neural network,'' \emph{arXiv preprint arXiv:1503.02531}, 2015.

\bibitem{55}\label{ref:55}
H. Yang, Y. Fang, W. Lin, and Z. Wang, ``Subjective quality assessment of screen content images,'' \emph{in Proc. IEEE International Workshop on Quality of Multimedia Experience}, pp. 257-262, Sept. 2014.

\bibitem{55a}\label{ref:55a}
T. Arici, S. Dikbas, and Y. Altunbasak, ``A histogram modification framework and its application for image contrast enhancement,'' \emph{IEEE Trans. Image Process.}, vol. 18, no. 9, pp. 1921-1935, Sep. 2009.

\bibitem{56}\label{ref:56}
I. Motoyoshi, S. Nishida, L. Sharan, and E. H. Adelson, ``Image statistics and the perception of surface qualities,'' \emph{Nature}, vol. 447, pp. 206-209, May 2007.
\end{thebibliography}
\end{document}